%% file: HRI2022.tex
\pdfoutput=1
\documentclass[sigconf]{acmart}

\usepackage{graphicx}
\usepackage{amsmath}
\usepackage{bm}
\usepackage[ruled,vlined,linesnumbered,noend]{algorithm2e}
\usepackage{algpseudocode}
\usepackage{hyperref}
\usepackage{xcolor}
\usepackage{balance}
\definecolor{Cerulean}{rgb}{0,0,0.95}
\definecolor{LimeGreen}{rgb}{0.15,0.65,0.15}
\definecolor{RoyalBlue}{rgb}{0.25,0.41,0.88}
\definecolor{Rose}{rgb}{1.0, 0.15, 0.21}
\definecolor{Orange}{rgb}{1.0, 0.5, 0.0}
\definecolor{Gray}{gray}{0.6}
\definecolor{Black}{gray}{0.0}
\definecolor{Purple}{rgb}{0.77,0.12,0.64}
\hypersetup{
    colorlinks=true,
    linkcolor=orange,
    filecolor=orange,      
    urlcolor=orange,
    citecolor=orange,
}

\AtBeginDocument{%
  \providecommand\BibTeX{{%
    \normalfont B\kern-0.5em{\scshape i\kern-0.25em b}\kern-0.8em\TeX}}}

\copyrightyear{2024}
\acmYear{2024}
\setcopyright{rightsretained}
\acmConference[HRI '24]{Proceedings of the 2024 ACM/IEEE International Conference on Human-Robot Interaction}{March 11--14, 2024}{Boulder, CO, USA}
\acmBooktitle{Proceedings of the 2024 ACM/IEEE International Conference on Human-Robot Interaction (HRI '24), March 11--14, 2024, Boulder, CO, USA}
\acmDOI{10.1145/3610977.3634970}
\acmISBN{979-8-4007-0322-5/24/03}

\AtBeginDocument{%
 \providecommand\BibTeX{{%
  Bib\TeX}}}

\makeatletter
\gdef\@copyrightpermission{
  \begin{minipage}{0.3\columnwidth}
   \href{https://creativecommons.org/licenses/by/4.0/}{\includegraphics[width=0.90\textwidth]{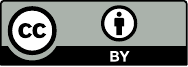}}
  \end{minipage}\hfill
  \begin{minipage}{0.7\columnwidth}
   \href{https://creativecommons.org/licenses/by/4.0/}{This work is licensed under a Creative Commons Attribution International 4.0 License.}
  \end{minipage}
  \vspace{5pt}
}
\makeatother

\begin{document}

\title{PREDILECT: Preferences Delineated with Zero-Shot Language-based Reasoning in Reinforcement Learning}

\author{Simon Holk}
\email{sholk@kth.se}
\orcid{0000-0001-5727-8140}
\authornote{These authors contributed equally}
\affiliation{%
  \institution{KTH Royal Institute of Technology}
  \country{Stockholm, Sweden}
}

\author{Daniel Marta}
\email{dlmarta@kth.se}
\orcid{0000-0002-3510-5481}
\authornotemark[1]
\affiliation{%
  \institution{KTH Royal Institute of Technology}
    \country{Stockholm, Sweden}
}

\author{Iolanda Leite}
\email{iolanda@kth.se}
\orcid{0000-0002-2212-4325}
\affiliation{%
  \institution{KTH Royal Institute of Technology}
\country{Stockholm, Sweden}
}

\renewcommand{\shortauthors}{Simon Holk, Daniel Marta, \& Iolanda Leite}
\input{content/settings}

\begin{abstract}
Preference-based reinforcement learning (RL) has emerged as a new field in robot learning, where humans play a pivotal role in shaping robot behavior by expressing preferences on different sequences of state-action pairs. However, formulating realistic policies for robots demands responses from humans to an extensive array of queries. In this work, we approach the sample-efficiency challenge by expanding the information collected per query to contain both preferences and optional text prompting. To accomplish this, we leverage the zero-shot capabilities of a large language model (LLM) to reason from the text provided by humans. To accommodate the additional query information, we reformulate the reward learning objectives to contain flexible highlights -- state-action pairs that contain relatively high information and are related to the features processed in a zero-shot fashion from a pretrained LLM. In both a simulated scenario and a user study, we reveal the effectiveness of our work by analyzing the feedback and its implications. Additionally, the collective feedback collected serves to train a robot on socially compliant trajectories in a simulated social navigation landscape. We provide video examples of the trained policies at \href{https://sites.google.com/view/rl-predilect}{https://sites.google.com/view/rl-predilect}

\end{abstract}


\maketitle
\input{content/01introduction}

\input{content/02relatedwork}

\input{content/03background}
\input{content/04method}
\input{content/06study}

\input{content/07conclusions}

\vspace{-5pt}
\begin{acks}
This research has been carried out as part of the Vinnova Competence Center for Trustworthy Edge Computing Systems and Applications at KTH, and partially supported by the Swedish Foundation for Strategic Research (SSF FFL18-0199) and the Wallenberg AI, Autonomous Systems  and  Software  Program (WASP) funded by the Knut and Alice Wallenberg Foundation. All of the authors are with the Division of Robotics, Perception and Learning, School of Electrical Engineering  and Computer Science, KTH Royal Institute of Technology, 114 28 Stockholm, Sweden. The authors are also affiliated with Digital Futures.
\end{acks}

\bibliographystyle{ACM-Reference-Format}
\balance
\bibliography{HRI2022}


\clearpage
\input{content/08appendix}

\end{document}

%% file: content/settings.tex
\newcommand{\trajectory}{\tau}
\newcommand{\highlight}{h}
\newcommand{\highlightdataset}{\mathcal{D}}
\newcommand{\highlightdiscount}{\lambda}
\newcommand{\highlightweight}{\alpha} 
\newcommand{\highlightlength}{L}
\newcommand{\segment}{\sigma}
\newcommand{\ntrajectories}{l}
\newcommand{\segmentdataset}{\mathcal{D}_{\sigma}}
\newcommand{\nsegments}{k}
\newcommand{\rewardnetwork}{\hat{\reward}_{\psi}}
\newcommand{\rewardnetworkobservation}{\rewardnetwork(s,a)}
\newcommand{\loss}{\mathcal{L}}
\newcommand{\dataset}{\mathcal{D}}
\newcommand{\preference}{w}
\newcommand{\query}{(\segment^0,\segment^1,\zeta)}
\newcommand{\datasettraj}{\mathcal{D_{\trajectory}}}
\newcommand{\datasetquery}{\mathcal{D}_q}
\newcommand{\preferencedataset}{\mathcal{D}}
\newcommand{\highlighteddataset}{\mathcal{D_{hq}}}

\newcommand{\stateseq}{\bm{s}_{\segment}}

\newcommand{\reward}{r}
\newcommand{\policy}{\pi}
\newcommand{\States}{\mathcal{S}}
\newcommand{\s}{s}
\newcommand{\Actions}{\mathcal{A}}
\newcommand{\action}{a}
\newcommand{\safeActions}{{\Actions_\mathrm{safe}}}

\newcommand{\naturalnumb}{\mathbb{N}_+}
\newcommand{\realnumb}{\mathbb{R}}

\newcommand{\ie}{i.e., }
\newcommand{\eg}{e.g., }

\newcommand{\feature}{f}
\newcommand{\prompt}{\text{prompt}}
\newcommand{\chatgpt}{\text{LLM}}
\newcommand{\response}{\text{r}_i}
\newcommand{\sentiment}{\text{y}}
\newcommand{\val}{\text{v}}
\newcommand{\mapping}{\text{M}}
\newcommand{\tensor}{\mathcal{T}}
\newcommand{\metric}{\text{m}}

%% file: content/01introduction.tex
\section{Introduction}

\begin{figure*}[h!]
	\centering
	\includegraphics[scale=0.75]{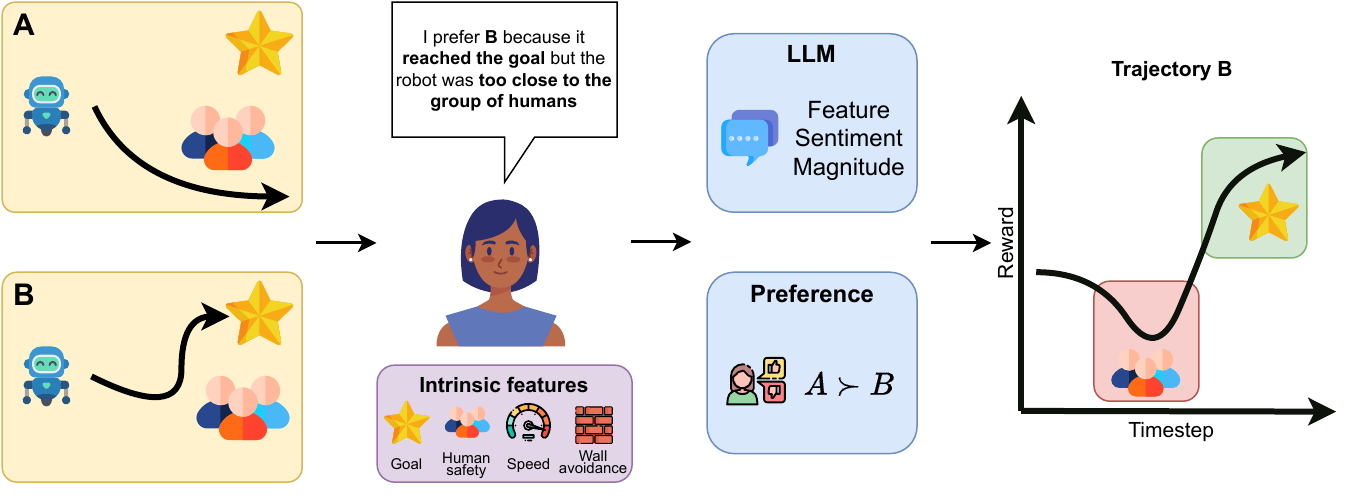}
	\caption{An overview of PREDILECT in a social navigation scenario: Initially, a human is shown two trajectories, A and B. They signal their preference for one of the trajectories and provide an additional text prompt to elaborate on their insights. Subsequently, an LLM can be employed for extracting feature sentiment, revealing the causal reasoning embedded in their text prompt, which is processed and mapped to a set of intrinsic values. Finally, both the preferences and the highlighted insights are utilized to more accurately define a reward function.}
	\label{fig:intro}
\end{figure*}

A key ingredient for the success of preference-based RL is that modern methods place minimal constraints on the modality of the reward function \cite{christiano2017deep,ibarz2018reward,wang2022skill,hejna2023few}, facilitating the formulation of complex objectives for robotic applications. While preference-driven teaching incorporates the essential element of structural alignment, vital for designing intricate objectives \cite{bobu2020less,jeon2020reward,booth2022revisiting}, its substantial reliance on extensive human feedback limits its applicability in real-world robotics \cite{liang2022reward,park2022surf,hu2022explaining,hejna2023few}. Moreover, while preferences do show a strong correlation with causality~\cite{spirtes2000causation,eberhardt2017introduction}, there is evidence indicating that preferences may not be sufficient as a standalone modality to thoroughly delineate the causal relationship among states, actions, and rewards. This challenge is identified as \emph{causal confusion}~\cite{de2019causal, tien2022study}. In a comprehensive empirical study on causal confusion in preference-based RL, \citet{tien2022study} highlighted that the introduction of spurious features and a rise in model capacity can induce causal confusion regarding the true reward function, even when learning from thousands of pairwise preferences. Overlooking this aspect can result in spurious correlations, which may ultimately lead to either reward exploitation \cite{amodei2016concrete,hadfield2017inverse} or the creation of a distributional shift~\cite{de2019causal}. Both scenarios necessitate additional interactions and can result in diminished performance, representing a failure in reward inference and leading to suboptimal behaviors.

We believe that a balance can be struck between the easiness of providing preferences \cite{christiano2017deep,hejna2023few} and offering an optional natural language interface for humans, in an effort to uncover the true causal relationship, thus greatly reducing the entropy of credit assignment, as natural language presents a more natural way for humans to interact \cite{khayrallah2015natural,li2021intention,pate2021natural}. While the integration of natural language can pose a significant challenge on its own, we leverage recent advancements in large pretrained foundational models~\cite{bommasani2021opportunities}, such as BERT\cite{devlin2018bert}, CLIP with GPT-2~\cite{mokady2021clipcap}, and GPT-3~\cite{brown2020language}. These models excel in various tasks, including text completion, image-text similarity, image captioning, and robot planning~\cite{zeng2022socratic}. Additional evidence indicates that adequately large language models possess the capability to execute complex reasoning~\cite{ho2022large,zeng2022socratic}, potentially revealing causal reasoning from human prompts and thereby mitigating aspects of causal confusion. For instance, \citet{wei2022chain} demonstrated that the generation of a thought chain---a sequence of intermediate reasoning steps--- enables the emergence of advanced reasoning abilities in sufficiently expansive language models.

To address the aforementioned limitations we leverage the zero-shot capabilities of pretrained models, we introduce \textbf{PREDILECT}: \textbf{PRE}ferences \textbf{D}el\textbf{I}neated with Zero-Shot \textbf{L}anguag\textbf{E}-based Reasoning in Reinfor\textbf{C}emen\textbf{T} Learning. Figure~\ref{fig:intro} provides a macroscopic depiction of our approach. Consider the two trajectories on the left side of the figure. While the queried person preferred trajectory B, when asked to justify their preference they mentioned that the robot being too close to the group of humans was unnecessary. In typical preference learning methods, where only the preferred trajectory is used as feedback, the result of this query could lead to causal confusion.  PREDILECT addresses the limitations inherent to preference-based RL by delineating preferences with highlights (sequences of state-action pairs) from the sentiment analysis of an LLM. The goal is to enhance the granularity of the reward model by partially uncovering the causal relationships between state-action and rewards. We achieve this by modifying the learning objective of the reward model with said highlights. We provide empirical evidence for the efficacy of PREDILECT, in the form of ablations on synthetic benchmarks. We also collect actual human preferences in a simulated social robot navigation scenario, to verify its applicability and further reinforce its merits and performance.

%% file: content/02relatedwork.tex
\section{Related Work}

\subsection{Learning from evaluative feedback}

Utilizing human knowledge for robot learning serves as an effective and interactive method \cite{macglashan2017interactive,zhang2022a}, particularly through the medium of evaluative feedback \cite{najar2021reinforcement}. By deducing reward functions from human input, we can facilitate the swift adaptation of robot policies \cite{xie2018few}, craft policies that are tailored to individual users \cite{shah2022offline}, and achieve alignment with instructions and descriptions \cite{sumers2022how}.
Humans can convey this form of feedback in a variety of manners, including scalar form \cite{knox2009interactively}, verbal directives \cite{sharma2022correcting}, trajectory segmentation \cite{cui2018active}, or by employing buttons to signal preferred behaviors \cite{knox2009interactively,knox2013training,senft2017supervised}, such as for expressing preferences \cite{wirth2017survey,christiano2017deep}. Prior studies have also focused on refining policies through preferences, either by pre-defining features \cite{cakmak2011human}, augmenting features~\cite{basu2018learning}, or utilizing Bayesian methods \cite{sadigh2017active,biyik2020active}.

Leveraging human preferences for learning has received substantial focus in recent literature \cite{wirth2017survey}, showcasing potential as an effective RL method applicable even in high-dimensional robotic settings \cite{hejna2023few}. Contemporary methods in preference learning impose few restrictions on reward function modality and can be interpreted as an iteration of repeated inverse reinforcement learning \cite{amin2017repeated}. A reward function can either be inferred from a \textit{tabula rasa} approach \cite{christiano2017deep} or be bootstrapped via imitation learning \cite{ibarz2018reward}. The underlying principle is the perpetual refinement of a reward function by soliciting preferences on subsequent iterations of a policy. Strategies that incrementally incorporate human participation in the learning loop to infer reward functions exhibit enhanced robustness and diminish the risk of obtaining contradictory feedback from humans \cite{christiano2017deep,baavenstrand2019performance,schrum2022mind}. Recent approaches have tackled the issue of feedback inefficiency utilizing techniques like pre-training \cite{lee2021pebble,park2022surf,hejna2023few} and bi-level optimization \cite{liu2022metarewardnet}. In contrast, PREDILECT delves into the promising and relatively uncharted challenge associated with preference explanation \cite{hu2022explaining}.

\subsection{Zero-shot multimodal prompting}
In this work, our aim is to integrate multimodal~\cite{ngiam2011multimodal} language information with preferences in a zero-shot fashion, leveraging large language models (LLMs)~\cite{devlin2018bert,brown2020language,thoppilan2022lamda,chen2021evaluating,radford2021learning}. LLMs are renowned for executing linguistic tasks, such as textual interpretation~\cite{rajpurkar2018know} and thus suitable for feature interpretation and extraction.  

Our interest lies in a variant of zero-shot transfer learning \cite{gavves2015active,schwab2018zero,ying2018transfer,soh2020meta}. Specifically, we seek to utilize the underlying reasoning found in human text prompts acquired with an LLM from a source domain (Internet-scale text prompts) with preferences, to both enhance performance and reduce the number of labeled queries needed in a target domain. During multimodal training~\cite{zeng2022socratic}, it is common to retain specific subcomponents of models—particularly those related to one modality but not others—in a frozen state for downstream tasks \cite{kulkarni2019unsupervised,florence2019self,zhai2021lit,tsimpoukelli2021multimodal,zakka2022xirl}. The fusion of weights from extensive pretrained models with multimodal joint training has led to notable accomplishments in diverse downstream multimodal applications, including image captioning~\cite{mokady2021clipcap}.

In alignment with our research, \citet{zeng2022socratic} presents Socratic Models (SMs), which are portrayed as modular frameworks. In these frameworks, new tasks emerge from a language-based interaction between pretrained models and additional modules, such as a reward model. PREDILECT can be conceptualized as an SM, where the large pretrained LLM is predetermined, while a reward model undergoes training based on joint text inferences between the LLM and preferences. Alternatively, our approach can be interpreted as a form of distillation \cite{radosavovic2018data,chen2020semi,xie2020self}, where the LLM serves as the teacher and effectively functions as a regularizer, aiding in the training of the student reward function. Inspired by the aforementioned works, PREDILECT contributes by utilizing the simplicity of preference-based RL and delving into the contextual text information associated with the respective queries.

%% file: content/03background.tex
\section{Background}
\label{sec:background}
We present the fundamentals to understand PREDILECT (see Sec.\ref{sec:predilect}). In this work, our goal is to develop a model that accurately predicts the reward function by efficiently leveraging human feedback, represented by preferences and prompts. The scenario we consider is one where a robot, in a given state \(s_t\), initiates an action \(a_t\) according to a policy \(\pi_\omega(a_t, s_t)\), parameterized by \(\omega\). Upon executing this action, the robot receives a reward \(r(s_t, a_t)\) and transitions to a new state \(s_{t+1}\), all within the framework of a Markov Decision Process (MDP). The final objective for the robot is to discover an optimal policy \(\pi_\omega^*(a_t, s_t)\) that maximizes the expected discounted sum of rewards.

\subsection{Preference Learning}
\label{sec:pref}

Following \citet{christiano2017deep}, we define the task of inferring a reward function, $\rewardnetwork$, characterized by $\psi$, from preferences as an issue in supervised learning. The core ambition of preference-based RL, as explored in~\cite{wirth2017survey,christiano2017deep}, revolves around deducing rewards from sequences of state-action pairs. Defining trajectory segments~\cite{wilson2012bayesian} as sequences constituted by state-action pairs, they are represented as $\segment^j = ((\s^j_t,\action^j_t),\dots,(\s^j_{t+m-1},\action^j_{t+m-1}))$, with $j$ signifying the segment index, encompassing state-action pairs from $t$ to $t+m$, where $m$ symbolizes the segment length. Humans are presented with pairs of these trajectory segments, designated as $(\segment^0,\segment^1)$, and are tasked with allocating a preference $\preference \in \{0,0.5,1\}$. A preference of $\preference=0$ signifies favoring $\segment^0$ over $\segment^1$, depicted as $\segment^0 \succ \segment^1$, while $\preference=1$ is interpreted as $\segment^1\! \succ \segment^0$, and $\preference=0.5$ indicates an equivalent preference for both segments. Adhering to the Bradley-Terry model~\cite{bradley1952rank}, the likelihood of a human exhibiting a preference for $\segment^0\! \succ \segment^1$, contingent upon it being exponentially reliant on the reward sum over the segments’ length, is expressed as:
\begin{equation}
\begin{split}
\!\!\!\!P_\psi[\segment^0\!\! \succ\!\! \segment^1] \!=\! \frac{\mathrm{exp}(\sum_{\substack{t}}\rewardnetwork(\s^0_t,\action^0_t))}{\mathrm{exp}(\sum_{\substack{t}}\rewardnetwork(\s^0_t,\action^0_t))\!+\!\mathrm{exp}(\sum_{\substack{t}}\rewardnetwork(\s^1_t,\action^1_t))}
\label{eq:softmax}
\end{split}
\end{equation}

In this framework, the reward model, $\rewardnetwork$, is amenable to training as a binary classifier to anticipate human preferences on new segments, serving as a surrogate for the reward function. The preferences elicited from humans are store alongside the respective segments in a labeled dataset $\dataset_l$, composed of triples $(\segment^0,\segment^1,\preference)$. During the optimization of $\rewardnetwork$, we draw samples from $\dataset_l$ and aim to minimize the binary cross-entropy loss:

\begin{equation}
\begin{aligned}
    \!\!\!\loss_{CE}(\rewardnetwork,\dataset_l) & = - \mathbb{E}_{\substack{(\segment^1,\segment^2,\preference) \sim \dataset_l }}  [(1\!\!-\!\!\preference) \;\text{log}\; P_\psi(\segment^1 \succ\segment^2) \\ & + \preference \;\text{log}\; P_\psi(\segment^2 \succ \segment^1)]
\end{aligned}
\label{eq:cross}
\end{equation}

%% file: content/04method.tex
\section{PREDILECT} 
\label{sec:predilect}

\subsection{Prompt-Response formulation}
\label{sec:prompt}

In PREDILECT a human can optionally offer a prompt to complement their preference, such that $\prompt_i\in\mathcal{P}$, where $i$ indexes the $i$-th prompt, and $\mathcal{P}$ the set of all prompts provided by the humans. 
To analyse human prompts, we require a set of intrinsic features $\mathcal{F} = \{\feature_1, \feature_2, \ldots, \feature_n\}$, where $n\in\naturalnumb$ represents the size of the feature set. It is important to note we do not want to bias humans on which features they should consider. Rather, we can denote intrinsic features we might find important for particular tasks, such as speed and distance to humans on social navigation. We leverage an LLM, to not only detect features from human prompts but provide sentiment analysis linked to these features. Thus, we input to an LLM both the prompt provided by the human and our feature set, such that:
\begin{equation}
    \chatgpt: (\prompt_i \in \mathcal{P}, \mathcal{F}) \rightarrow \response\in\mathcal{R}
    \label{eq:chatgpt}
\end{equation}
where $\mathcal{R}$ corresponds to the set of all possible responses. Each $\response$ consists of a set of triplets of the size $n=|\mathcal{F}|=|\mathcal{R}|$, where $\response = \{ (\feature_1, \sentiment_1, \val_1), \ldots, (\feature_n, \sentiment_n, \val_n) \}$, where $\feature_1\in\mathcal{F}$ is a feature, $\sentiment_1\in\{\text{positive},\text{negative}\}$ is the sentiment associated with feature $\feature_1$, and $\val_1\in\{\text{low},\text{high}\}$ is the magnitude associated with the sentiment $\sentiment_1$. To clarify our formulation we provide a concrete example of a prompt that was evaluated for our experiments (see Sec.~\ref{sec:experiments}) in the social navigation environment. For visualisation purposes, the feature set $\mathcal{F}$ is \textcolor{RoyalBlue}{blue}, the human prompt $\prompt_i$ is \textcolor{Purple}{purple} and the output response $\response$ is \textcolor{LimeGreen}{green.}:
\newline
\newline
\noindent\fbox{%
    \parbox{0.97\linewidth}{%
        \scriptsize{\texttt{{\color{Orange} Input: \color{Gray}You are a robot navigating a corridor with humans walking around trying to reach the goal/star. The user had to pick between two alternatives and picked their preferred alternative and they are now giving an explanation for their pick. Which feature(s) was most important of  {\color{RoyalBlue}[distance to goal, distance to human, speed]}? The text given by the user is: \color{Purple}'was less close to hitting a human/wall and moved at a slower pace.'\color{Gray} Please respond in the following format for each feature that is relevant to the text given by the user: [feature:insert feature, sentiment:insert positive or negative, value: insert high or low]. Sentiment explains if the user thought the robot was behaving well in regards to the feature, if the robot behaved well it should be positive, else negative. Value indicates if the value of the feature was high or low. Only mention the features that are relevant, disregard the others. \newline \color{Orange} Output:\newline {\color{LimeGreen}[feature: distance to human, sentiment: positive, value: high]\newline[feature: speed, sentiment: positive, value: low]}}}}
    }%
}\\
\vspace{-10pt}
\subsection{Mapping responses to highlights}
\label{sec:highlights}

We use each response $\response$ to obtain highlighted subsequences from a trajectory segment $\segment$, which we define as \emph{highlights}. A highlight, denoted as $\highlight$, is characterized as a subsequence residing within a preferred segment. 

Highlights are constructed as subsequences of segments, represented by $\highlight=\segment_{i,j}=((\s_i,\action_i),\dots,(\s_j,\action_j))\in(\s, \action)^{j-i},i,j\in\naturalnumb$ with $0\leq i \leq j \leq m$. The highlight’s length is given by $j-i=L$, wherein $L$ signifies the maximum length contemplated for a highlight. To map each $\response$ to highlights, we first semantically map each trajectory segment $\segment$ to a tensor of feature metrics $\tensor$. 

We define a mapping function such as $\mapping: \segment \rightarrow \tensor$, that maps the corresponding state sequence to a tensor of feature metrics $\tensor$. Each $\tensor_{ij}$ is the tensor element corresponding to the $i^{th}$ state and $j^{th}$ feature. Moreoever, each tensor element can be defined as $\tensor_{ij} = \{ \feature_j, \metric_{ij}\}$ which contains a feature $\feature_j$ and its corresponding metric value $\metric_{ij}\in\realnumb$ such that the tensor of feature metrics can be defined as: 
  \begin{equation}
\tensor = 
\begin{bmatrix}
\{ \feature_1, \metric_{11} \} & \{ \feature_2, \metric_{12} \} & \ldots & \{ \feature_n, \metric_{1n} \} \\
\{ \feature_1, \metric_{21} \} & \{ \feature_2, \metric_{22} \} & \ldots & \{ \feature_n, \metric_{2n} \} \\
\vdots & \vdots & \ddots & \vdots \\
\{ \feature_1, \metric_{m1} \} & \{ \feature_2, \metric_{m2} \} & \ldots & \{ \feature_n, \metric_{mn} \} \\
\end{bmatrix}
\end{equation}
In this context, each $\metric_{ij}$ is a scalar derived through a heuristic, which is intrinsically associated with the relevant feature. For instance, "distance to humans" is expected to represent a scalar value, measured in meters.

We require a search function $g$ that navigates through the tensor of metrics $\tensor$ to identify a highlight $h$ for each $\response$. Given a response $\response$, our task is to pinpoint, for every feature $f \in F$ referenced in $r_i$, a singular subsequence $S_{f}^{'} \subseteq S$ that aligns with the specified criteria. The set of all possible subsequences of fixed length $L$ is represented as ${ S' \subseteq S : |S'| = L }$, and from this set, for each feature $\feature$, we extract one subsequence to form the set $\mathcal{H}_{\mathcal{F}}(S) = \{ S^{'}_f \subseteq S : |S^{'}_f| = L \mid f \in \mathcal{F} \}$. The search function $g$ is defined as follows:
\begin{equation}
g(\mathcal{T}, R) = \mathcal{H}_{\mathcal{F}}(S)
\end{equation}
Here, $\mathcal{H}_{\mathcal{F}}(S)$ denotes the set containing a unique subsequence for each feature in $\mathcal{F}$. The function $g$ maps the tensor of metrics $\mathcal{T}$ and the response set $R$ to $\mathcal{H}_{\mathcal{F}}(S)$, aligning each subsequence with the user-specified prompt for the corresponding feature. For clarity, we will refer to the unique subsequence for each feature $f$, denoted as $S'_f$, as the \emph{highlight} $\highlight_i$ (as aforementioned defined) for feature $\feature_i$.

In PREDILECT, we elucidate further on the formulation of $g$. For each triplet $(\feature_k, \sentiment_k, \val_k)$ in $\response$, $\val_k$ is utilized to probe for either low or high metric values in $\tensor$, with $\sentiment_k$ delineating whether the identified highlight is of a positive or negative nature. Consequently, $g$ is articulated to search for highlights corresponding to either the maximum or minimum metric values, contingent on $\val_k$, and subsequently categorizes them based on the sentiment $\sentiment_k$ as positive or negative highlights. Subsequently, we bifurcate $\mathcal{H}_{\mathcal{F}}(S)$ into two distinct subsets, denoted as $\mathcal{H}_{\mathcal{F}}^+$ and $\mathcal{H}_{\mathcal{F}}^-$, based on the sentiment of each feature.

\begin{figure}[ht]
	\centering
	\includegraphics[clip, trim=0.6cm 0cm 0.4cm 0cm,  width=1\linewidth]{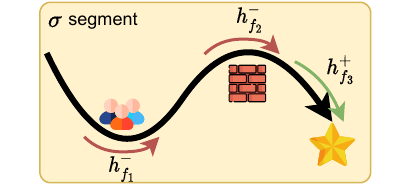}
	\caption{Representation of highlights within a segment. The segment $\segment$ outlined by a curve contains multiple highlights, two negative in $\mathcal{H}_{\mathcal{F}}^-$ depicted in red and one positive $\mathcal{H}_{\mathcal{F}}^+$ depicted in green $\highlight^+$. All highlights are of the same length $L$.}
	\label{fig:segment}
\end{figure}

A query, as delineated in Sec.~\ref{sec:pref}, encapsulates a human preference, symbolized as $\preference$, and is paired with two trajectory segments, represented as $q=(\segment_1, \segment_2, \preference)$. This definition is augmented by incorporating sets of positive and negative highlights, $\mathcal{H}_{\mathcal{F}}^+ = \{\highlight_{f_1}^+,\dots,\highlight_{f_n}^+\}$ and $\mathcal{H}_{\mathcal{F}}^- =\{\highlight_{f_1}^-,\dots,\highlight_{f_n}^-\}$, respectively, for each feature. The enhanced quintuple, termed as sentiment highlighted query, is denoted as $shq=(\segment_1, \segment_2, \preference,\mathcal{H}_{\mathcal{F}}^+, \mathcal{H}_{\mathcal{F}}^-)$ and we provide a visualization of these highlights in Fig.~\ref{fig:segment}. Within PREDILECT, the sentiment highlighted queries are compiled in a dataset represented as $\highlightdataset_{shq}$.

\begin{figure*}[ht]
	\centering
	\includegraphics[clip, trim=0.1cm 0.2cm 0.3cm 0cm,  width=1\linewidth]{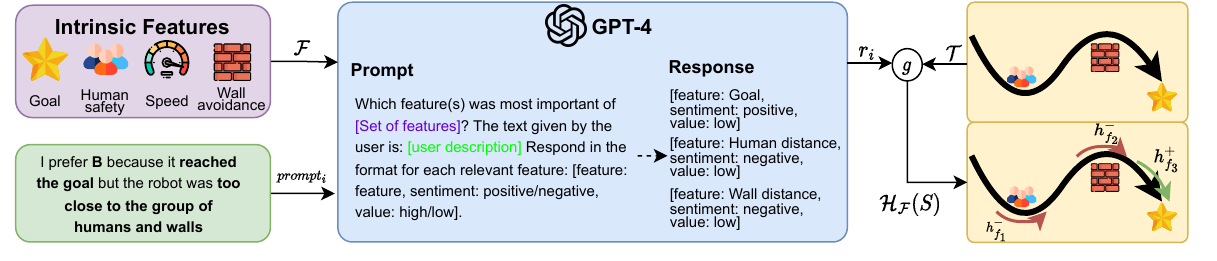}
	\caption{An overview of how PREDILECT processes prompts from humans is as follows: Initially, a human provides a prompt, depicted in \textcolor{LimeGreen}{green}, along with a set of intrinsic features $\mathcal{F}$ in \textcolor{Purple}{purple} which is environment dependant. Both are input into the LLM (ChatGPT-4 in the case of PREDILECT) to generate a response $\response$. Subsequently, after mapping a segment $\segment$ to a tensor of metrics $\tensor$ using the mapping function $M$, we apply a searching function $g$ to obtain the set $\mathcal{H}_\mathcal{F}$ of highlights for each feature. These highlights are then utilized to train our reward model $\rewardnetwork$ as per Eq.\ref{eq:predilect}.
}
	\label{fig:framework2}
\end{figure*}

\subsection{Reward Model Regularization}
\label{sec:regulamos}

Empirical studies illustrate the importance of strategic regularization in shaping state representations by enhancing the initial learning objective \cite{de2018integrating}. Auxiliary tasks, secondary yet semi-related to the primary task, offer valuable training signals for learning shared representations, thereby enhancing learning and data efficiency \cite{jaderberg2016reinforcement, mirowski2016learning, matas2018sim, shelhamer2016loss,pinto2017learning}. Regularization limits the search for solutions by adding bias. Using human natural language for shaping representations develops human-like biases and behaviors \cite{bhatia2020preference,kumar2022using}. In PREDILECT, we devise a state representation task incorporating causal reasoning from human teachers to reduce the entropy of the credit assignment when we only use preferences, refining the distinction between high and low-value sequences. This task is added as additional regularization terms to Eq.\ref{eq:cross} to shape the reward function, aiming to maximize positive highlights $\highlight^+$ and minimize negative ones $\highlight^-$. A discount is applied to preceding states, as in similar works predicting future rewards.
\begin{equation} \label{eq:positiveloss}
\loss_{+} = \mathbb{E}_{\highlight^{+} \sim \highlightdataset_{shq}} \left[\sum^{L}_{l=0} \highlightdiscount^{l}\rewardnetwork(s_{j-l}, a_{j-l})  \right]
\end{equation}

\begin{equation} \label{eq:negativeloss}
\loss_{-} = - \mathbb{E}_{\highlight^{-} \sim \highlightdataset_{shq}} \left[\sum^{L}_{l=0} \highlightdiscount^{l}\rewardnetwork(s_{j-l}, a_{j-l}) \right]
\end{equation}
The final objective optimizes both $\loss_+$ (Eq.~\ref{eq:positiveloss}) and $\loss_-$ (Eq.~\ref{eq:negativeloss}) while sustaining the baseline preference learning loss, denoted as $\loss_{CE}$ (see Eq.\ref{eq:cross}). To optimize $\rewardnetwork$, we utilize sentiment highlighted query samples $shq$ drawn from $\highlightdataset_{shq}$. The hyperparameters $\highlightweight_+$ and $\highlightweight_-$ are employed to assign weights to the two regularization terms. Consequently, the resulting learning objective takes the form:

\begin{equation}
\label{eq:predilect}
\loss_{\text{PREDILECT}} = \loss_{CE} + \highlightweight_+ \loss_{+} + \highlightweight_- \loss_{-}
\end{equation}

\subsection{Preference learning with PREDILECT}
\label{sec:algopredilect}

\begin{figure}[h]
        \centering
	\includegraphics[clip, trim=0cm 0.6cm 0.8cm 0cm, width=0.94\linewidth]{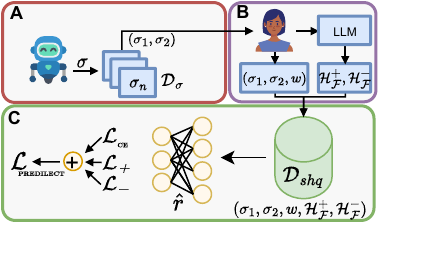}
	\caption{Framework representation of PREDILECT. Step A: We train policy $\policy_\omega$ and sample rollouts which are stored in $\segmentdataset$. Step B: We sample trajectory segments $\segment$ to query humans and collect both preferences and prompts. The prompts are processed through an LLM to obtain responses. Those responses are used to obtain highlights $(\mathcal{H}_\mathcal{F}^+, \mathcal{H}_\mathcal{F}^-)$ from the preferred segment $\segment^*$ Step C: The sentiment highlighted queries are collected to form dataset $\highlightdataset_{shq}$ and update the current reward model $\rewardnetwork$.} 
	\label{fig:frameworkPREDILECT}
	\vspace{-20pt}
\end{figure}

Similar to other preference-based RL methods \cite{christiano2017deep,lee2021pebble}, PREDILECT, as delineated in Alg.~\ref{algo:predilect}, integrates policy with reward learning. In step A (see Fig.~\ref{fig:frameworkPREDILECT}), policy $\policy_\omega$ engages with the environment, yielding $(s_t,a_t,s_{t+1})$ and estimates of $\rewardnetwork(s_t,a_t)$. The resulting transitions $(s_t,a_t,s_{t+1},\rewardnetwork(s_t,a_t))$, organized into trajectories, are gathered in a temporary buffer. This buffer is utilized for gradient descent on $\policy_\omega$ concerning $\omega$, following the PPO algorithm \cite{schulman2017proximal}. Post training of $\policy_\omega$, a substantial number of trajectory segments are collected and stored in $\segmentdataset$.
Subsequently, in step B, a feedback session takes place. We randomly select $N \in \naturalnumb$ trajectory segments $\segment$ to obtain query preferences from humans. In addition, they can provide optional auxiliary language feedback. Given the human prompt and the set of intrinsic features, we utilize the LLM to derive a response $\text{r}$. After transforming the segment $\segment$ into a tensor of metrics $\tensor$ using $M$, we employ the search function $g$ to extract both positive $\mathcal{H}_\mathcal{F}^+$ and negative highlights $\mathcal{H}_\mathcal{F}^-$ from segments and prompts, as described in Sec.~\ref{sec:highlights}. All highlights are subsequently stored in the sentiment highlighted queries dataset $\highlightdataset_{shq}$.

In step C, gradient descent is executed on the parameters $\psi$ to refine our reward model $\rewardnetwork$, with $\loss_{\text{PREDILECT}}$ serving as the learning objective. Upon acquiring an updated version of $\rewardnetwork$, the process reverts to step A, and the algorithm iteratively progresses until convergence.

\begin{algorithm}

\SetAlgoLined

 $\preferencedataset_{\segment} \leftarrow \emptyset;$ 
 $\preferencedataset_{shq} \leftarrow \emptyset;$ 
$\mathcal{F} \leftarrow \mathrm{getFeatureSet();}$ \\

 $\policy_\omega \leftarrow \mathrm{train}(\policy_\omega,\rewardnetwork,env)$  \Comment{\tcp{Step A}}
 
 $\segmentdataset\leftarrow \mathrm{sampleSegments(\policy_\omega)}$ \\

\While{$|\preferencedataset_{shq}| \leq N$}
{
 $(\segment_1,\segment_2) \leftarrow \mathrm{samplePairs(\segmentdataset)}$ \Comment{\tcp{Step B}}
 $\preference \leftarrow \mathrm{collectPreference(\segment_1,\segment_2)}$ \\
 $ \prompt \leftarrow \mathrm{collectPrompt(\segment_1,\segment_2, \preference)}$\\
 $ \text{r} \leftarrow \text{LM}(\prompt,\mathcal{F})$\\
 $ \tensor \leftarrow \mathrm{M(\segment^*)}$\\
 $(\mathcal{H}_\mathcal{F}^+, \mathcal{H}_\mathcal{F}^-) \leftarrow \mathrm{g(\tensor,\text{r})}$\\
  $ \preferencedataset_{shq} \leftarrow  \preferencedataset_{shq} \cup (\segment_1,\segment_2,\preference, \mathcal{H}_\mathcal{F}^+, \mathcal{H}_\mathcal{F}^-)$
}

 \For{each gradient step}{
 Sample minibatch from $\preferencedataset_{shq}$ \Comment{\tcp{Step C}}
 Optimize $\rewardnetwork$ $\loss_{\textbf{PREDILECT}}$ with respect to $\psi$ \text{in} Eq.\text{(\ref{eq:predilect})} 
 }

  \textbf{return} $\rewardnetwork$ \Comment{\tcp{return $\rewardnetwork$}}

 \caption{PREDILECT}\label{algo:predilect} 
\end{algorithm}

\label{sec:algorithm}

%% file: content/06study.tex
\section{Experiments}
\label{sec:experiments}
This section aims to examine the efficiency of PREDILECT and how incorporating human language to express preferences impacts the preference learning framework. To begin, we conduct experiments in simulated settings. We create an oracle that offers feedback to demonstrate how merging preference learning with textual explanations can enhance outcomes. We also carry out an experiment in a social robot navigation context, where a robot is trained using real human feedback sourced from Amazon Mechanical Turk (MTurk) participants. Social robot navigation is a complex task, requiring a balance between objectives such as reaching the destination, efficiency, and (perceived human) safety, making it a compelling case study~\cite{francis2023principles}. We assess the effectiveness of PREDILECT with human feedback and how well the LLM can capture essential information from these textual descriptions to produce highlights. We used GPT-4 for all experiments. We also demonstrate our ability to develop varied policies, such as those prioritizing safety, by only considering features related to safety when training the human reward function. Our main hypotheses are summarized below:

\begin{itemize}
  \item H1: PREDILECT will learn a human reward function more efficiently compared to the baseline.
  \item H2: PREDILECT can learn policies that put more focus on specific objectives as described by the user rather than more generalized policies learned by regular preference-based learning.
  \item H3: The LLM will accurately extract the information needed to create highlights.
\end{itemize}

\subsection{Simulation experiment setup}
We aim to demonstrate the effectiveness of PREDILECT by conducting simulated experiments using the Reacher and Cheetah environments from OpenAI Gym \cite{brockman2016openai}. In these simulations, an oracle compares two segments and provides preferences based on which segment achieves a higher cumulative reward, as determined by the true environment reward function. Recognizing that human instructors are not infallible, we introduce a 10\% error rate in the oracle's feedback to mimic potential human inaccuracies.

We've also extended the oracle to work with PREDILECT. Besides just indicating a preference, the oracle also offers explanations for its choices. It does this by monitoring the values of certain features within a segment. If a feature's value surpasses or falls below a set threshold, the oracle will add this feature for highlighting. This can be seen as PREDILECT after the first LLM processing step. The specific features that the Oracle monitors vary by environment. For Cheetah, we use the x-axis velocity; for Reacher, we use the distance between the fingers and the goal. Further details can be found in the Appendix.

\subsection{Simulation experiment results}
\begin{figure*}[h]
	\centering
	\includegraphics[clip, trim=0cm 0.2cm 0cm 0.2cm,width=0.95\linewidth]{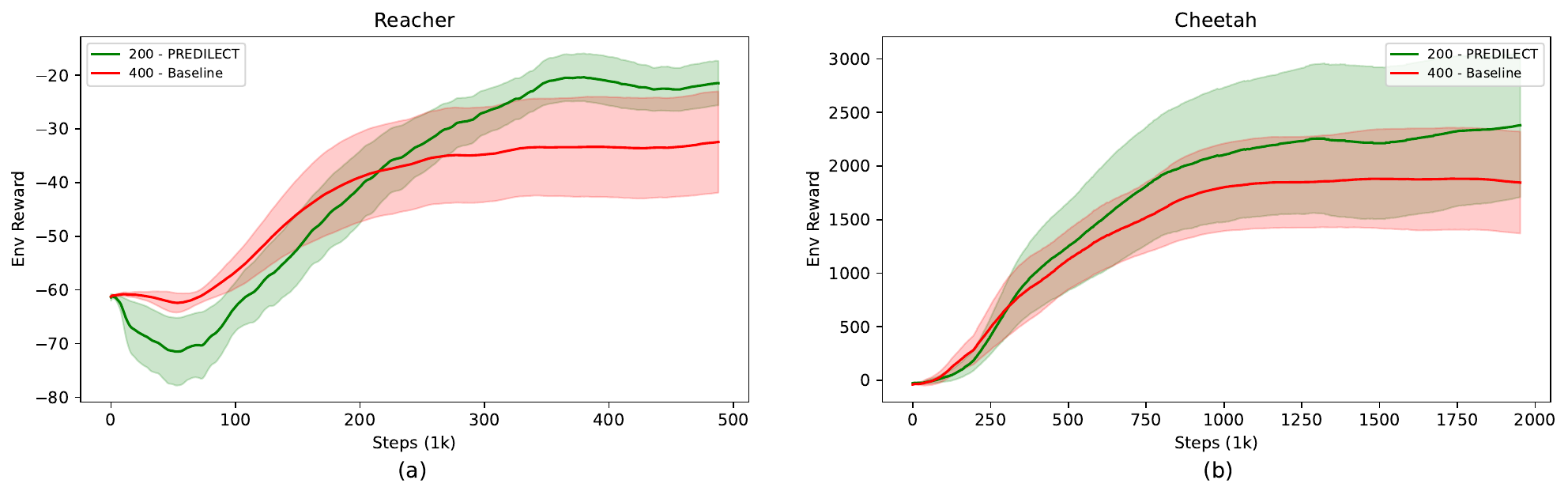}
	\caption{Learning curves: for Reacher (a), PREDILECT used 200 queries, Baseline used 400 queries; for Cheetah (b), PREDILECT used 200 queries, Baseline used 400 queries.}
	\label{fig:simcurves}
	\vspace{-5pt}
\end{figure*}

\begin{table}
\begin{center}
\resizebox{0.45\textwidth}{!}{\begin{tabular}{p{2.5cm}p{2cm}p{2cm}}
    \hline
     PREDILECT 200 & LLM 200 & LLM 400 \\
    \hline
    2378  & 635 & 968  \\

    \hline
  \end{tabular}}
   \caption{Ablation comparing the final cumulative reward when using only $\loss_{+}$, $\loss_{-}$ and PREDILECT for Cheetah.}
  \label{tab:llm-only-ablation}
\vspace{-20pt}
\end{center}
\end{table}

Upon initial observation, it's evident that utilizing highlights based on features results in quicker convergence compared to relying solely on preference-based learning (see Fig \ref{fig:simcurves}). The results derive from highlights pinpointing more specific and non-sequential areas of interest that align with the provided description. This advantage is evident in the reward curves for both Cheetah and Reacher. Notably, this enhanced performance with PREDILECT is achieved using only half the number of queries typically required by traditional preference-based learning. Tab.~\ref{tab:llm-only-ablation} further demonstrates that the high convergence stems from the multi-modal feedback. The simulated results offer support for H1 due to the higher convergence and reduction of queries.

\subsection{Real human feedback experiment setup}
To understand how textual descriptions affect the learned reward model, we perform experimentation using real human feedback in a social navigation scenario. The purpose of this experiment is threefold. 1) validate that PREDILECT performs better than baseline using real human feedback; 2) show that the policies learned can be more aligned with the participant's preferences; 3) show that the LLM can accurately deduce the information needed from the textual descriptions.

\begin{figure}
    \centering
\includegraphics[width=1\linewidth, trim=0 0 0 10, clip]{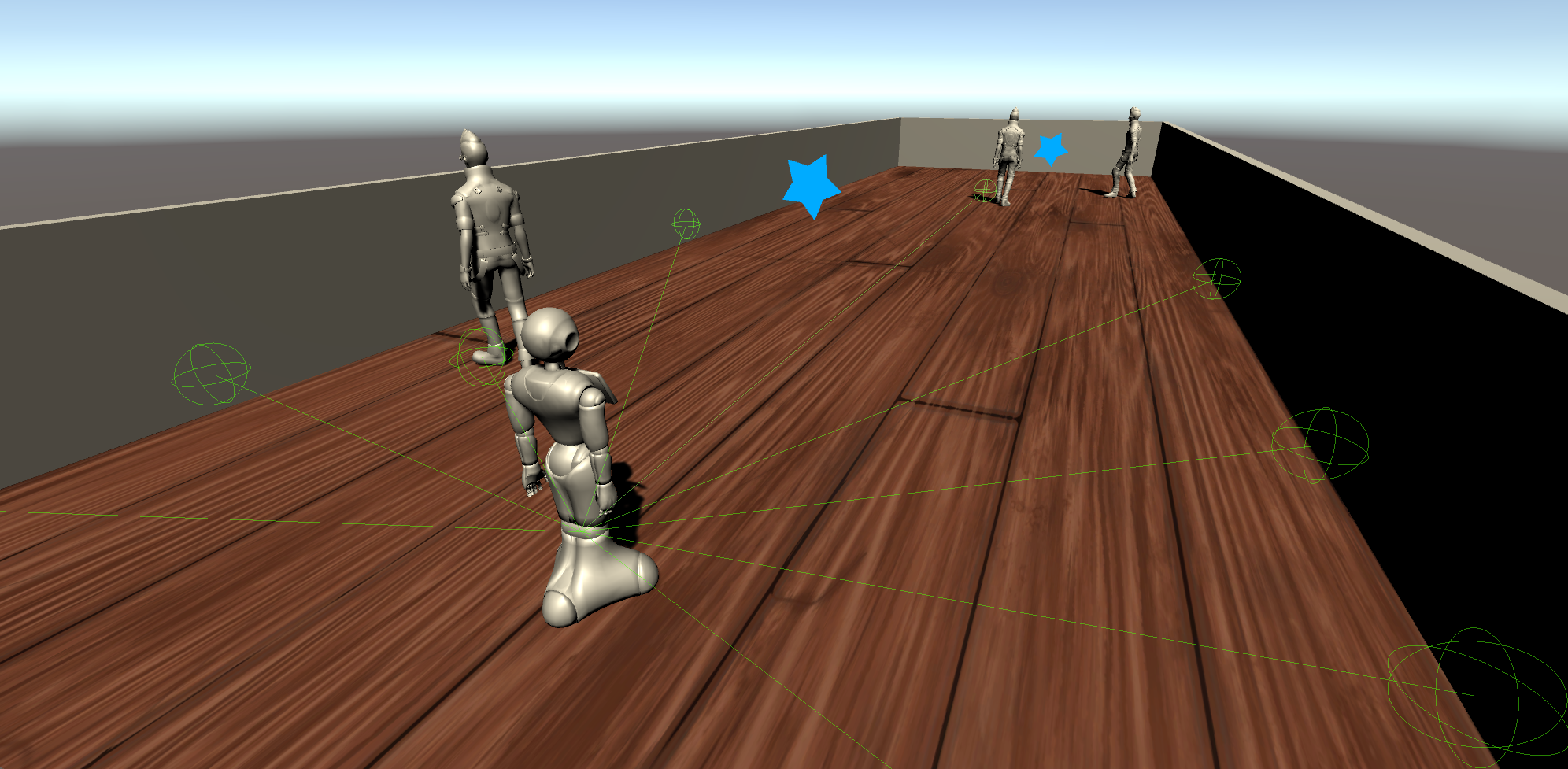}
    \caption{Social robot navigation environment used for the human experiments.}
    \label{fig:socialnavigation}
    \vspace{-5pt}
\end{figure}

The social navigation scenario is built using Unity and involves a Pepper robot navigating between three humans in order to collect a goal shaped like a star (see Fig \ref{fig:socialnavigation})that acts as a guide for the robot. In order to sense its environment the robot is equipped with lidar rays that can detect humans, walls, and the end goal. To ensure safety, the robot follows the social force model which treats the human and robot as repelling forces~\cite{Helbing_1995}. One of the actions the robot can take is to lower and increase the social force which will compel the robot to keep its distance from the humans. For PREDILECT we add the features for goal distance, human distance, and speed to the prompt. The agent runs for 500,000 timesteps and updates the reward function once at the start as a single batch. 
\vspace{-6pt}
\subsubsection{Participants} In total, 43 individuals were recruited from Amazon Mechanical Turk to participate in the experiment: 21 were assigned to the PREDILECT group and 22 to the baseline group. However, three participants were excluded due to not passing an attention check, resulting in an even 20 participants for each group. The participants' ages ranged from 22 to 59 years. Of them, 64.9\% identified as male and 35.1\% as female. The experiment lasted about 25 minutes and participant were compensated at a rate of approximately \$10 per hour for their time.
\vspace{-6pt}
\subsubsection{Procedure} After agreeing to participate in the experiment, participants went through a tutorial explaining the goal and the UI. They were instructed to embody the humans in the environment and rate the queries based on their comfort and how well the robot achieved its goal. Participants were then provided with pairs of videos that represent two segments collected by the robot following a pre-trained policy $\policy_\omega$ forming a query, and had to indicate which video they preferred. If they couldn't decide which video they preferred, they had the option to say "none". After each preference, they were moved to the next query pair until they had labeled 20 queries in total. At the end of the experiment, the participants are asked to fill in a demographic survey.

In the PREDILECT trial, before proceeding to the next query, participants were prompted with a question: "Please describe why you preferred video A/B." They were provided with a text box to record their response. If they were uncertain about their preference reasons, they were instructed to leave the text box blank. To assist them in recalling the video's content, the selected video remained centered on the screen, allowing participants to rewatch it instead of relying solely on memory. Table \ref{tab:descriptions} includes some examples of textual descriptions given by participants and how the LLM module in PREDILECT mapped these into features.  

We conducted separate trials for both PREDILECT and the baseline, gathering 400 preferences for each. Both trials began with the same initial set of segments and preference pairs, all generated from the same policy $\policy_\omega$. The distinctions between the two arise from how they evaluate preference pairs and the inclusion of textual descriptions in PREDILECT. 
\subsection{Real human feedback experiment results}
\begin{figure*}[h]
	\centering
	\includegraphics[clip, trim=0cm 0.2cm 0cm 0.2cm,width=0.95\linewidth]{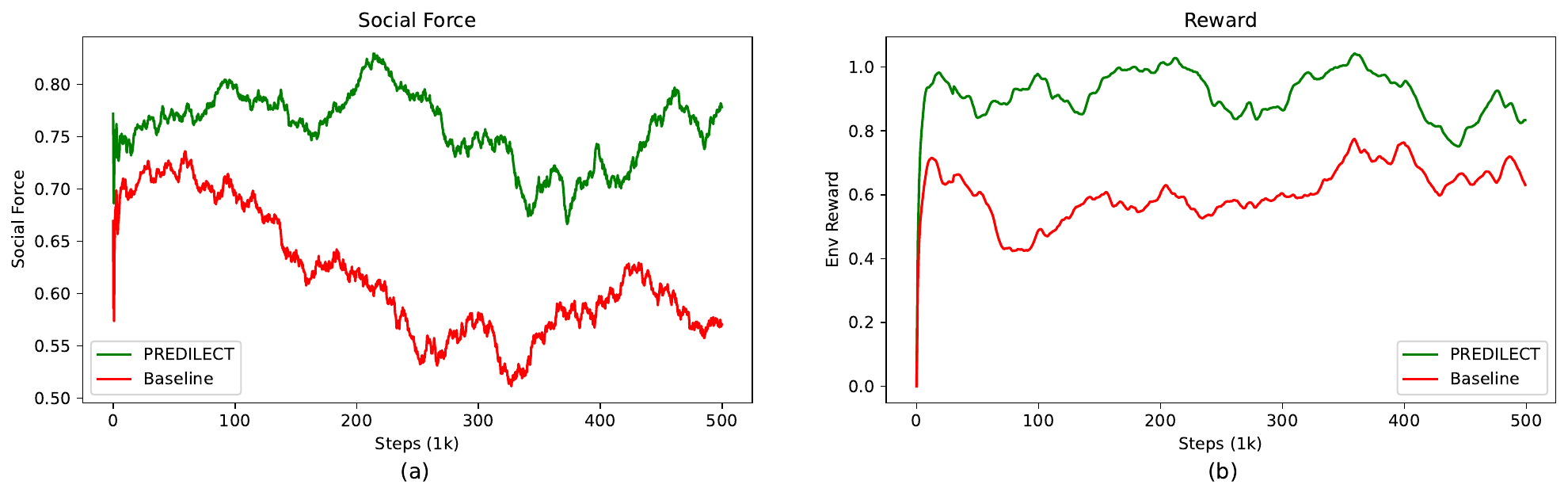}
	\caption{For the social robot navigation environment: a) Social force when applying only safety features on PREDILECT compared to baseline; b) Reward curves for PREDILECT and baseline.}
	\label{fig:socialnav}
	\vspace{-5pt}
\end{figure*}

We further assess the algorithm's efficiency by using actual human feedback. When comparing PREDILECT to baseline preference learning, PREDILECT converges to a higher reward and stays relatively stable at just above 0.8 reward while the baseline converges around 0.6, as shown in Figure \ref{fig:socialnav}.b. These results add further support for Hypothesis H1 on top of the simulation results. Given that humans can further articulate their preferences using text, we believe it's beneficial to investigate how we can tailor policies based on these textual descriptions in order to validate our hypothesis H2. With this in mind, we use PREDILECT while focusing solely on safety-related features, such as the proximity to humans and walls. Figure \ref{fig:socialnav}.a illustrates a distinct difference in the robot's application of social force when trained using only these safety-centric features compared to baseline. Initially, both the baseline and our method exhibit similar values. However, over time, the baseline value diminishes, which might be attributed to a trade-off between maintaining a safe distance and efficiently reaching the destination. The ability to have the robot use a higher social force by training using safety-based description offers support for H2.
To ensure that the LLM effectively and accurately extracts information from human descriptions and to validate hypothesis H3, we compared its predictions with those of a human labeler, as detailed in Table \ref{tab:llmmetrics}. The data reveals that the LLM correctly identifies features from the descriptions 85\% of the time. It also has similar accuracy rates for determining sentiment ($\sim$77\%) and magnitude ($\sim$80\%).

While these accuracy rates are noteworthy, they also encompass instances where the LLM might overlook certain features that the human labeler identified. In such cases, we lose some information and revert to standard preferences. However, a more important concern arises when the LLM identifies or "hallucinates" features that the human never mentioned, as this could adversely impact performance. Our data shows that the LLM makes this error 11.54\% of the time when identifying features. Furthermore, in situations where both the human labeler and the LLM agree on a particular feature, there's a discrepancy in sentiment interpretation in 13\% of the instances and in magnitude interpretation in 8\% of the instances. This error rate is based solely on the human descriptions. Interestingly, when considering our prompt we have a general goal description of the robot along with the human language description. When considering the goal description, most of the cases where there is an error, they are still aligned with the overarching goal description. With these accuracies and the fact that any error can mostly be explained with the prompts structure, we find support for H3. Overall the participants chose the "none" alternative 13.75\% of the time, indicating no preference. For PREDILECT when there was a preference, the participants gave additional textual feedback in ~61.44\% of the cases.

\begin{table}
\begin{center}
\resizebox{0.40\textwidth}{!}{\begin{tabular}{p{0.5cm}p{1.3cm}p{1.3cm}p{1.3cm}}
    \hline
    \multicolumn{4}{c}{\textbf{LLM metrics}}\\
    & Feature & Sentiment & Magnitude \\
    \hline
    Acc & 85.71\%  & 77.14\% &  80\%  \\
    F.pos & 11.54\%  & 13.04\% &  8\% \\

    \hline
  \end{tabular}}
   \caption{Accuracy and false positives of the LLM is at predicting the feature, sentiment, and magnitude based on human description when compared to a human labeler.}
  \label{tab:llmmetrics}
\end{center}
\vspace{-15pt}
\end{table}

\begin{table}
\begin{center}
\resizebox{0.45\textwidth}{!}{\begin{tabular}{ |p{3cm}||p{1.2cm}|p{1.2cm}|p{1cm}|  }
 \hline
 Text Description & Feat. & Sent. & Magn. \\
 \hline
\textit{got both stars and avoided hitting humans }  & H.Dist, G.Dist & Positive, Positive & High, Low\\
 \hline
\textit{ I prefer video A because of the fast speed.}& Speed  & Positive & High\\
\hline
\textit{Did not hit a human} &H.Dist & Positive & High\\
 \hline
\textit{Stayed a longer distance away from hitting anything and moved at a slower pace. }  &H.Dist, Speed & Positive, Negative &  High, High\\
 \hline
\end{tabular}}
\caption{Human descriptions justifying their preferences and the feature, sentiment, and magnitude predicted by the LLM.}
\label{tab:descriptions}
\vspace{-20pt}
\end{center}
\end{table}

%% file: content/07conclusions.tex
 \vspace{-3pt}
\section{Discussion}

Based on the results from both the simulated oracle experiments and the social robot navigation evaluations using real human feedback, PREDILECT demonstrates a faster convergence compared to traditional preference learning. In the simulated environments, this higher efficiency is obtained using 50\% of the queries compared to the baseline. This superior performance can be attributed to the feedback's capacity to mitigate causal confusion, as it enables humans to provide a clear rationale behind their preferences.

Our results further illustrate by emphasizing particular features, such as those related to safety, our system can tailor policies based on user explanations rather than merely adopting a generic policy. For instance, within the social navigation context, the robot consistently maintained a higher level of social force during training. In contrast, the baseline reduced the social force over time. We contend that the integration of textual explanations for preferences allows the system to more rapidly align with human preferences, eliminating the need for over-querying making this approach relevant to several robotics applications.

We evaluated the LLM's capability to extract relevant information from text descriptions provided by humans. The LLM demonstrated high accuracy, closely aligned with the feature identification by a human labeler. While there were occasional discrepancies, a detailed analysis revealed that the LLM's interpretations, even when diverging from participants' descriptions, frequently aligned with the robot's overarching objective. This may be attributed to the context provided within our prompt, which briefly described the robot's task. It is noteworthy that the LLM considers the entirety of the context presented in the prompt unless explicitly directed otherwise. This observation is insightful for two reasons: it underscores the importance of crafting prompts, and it highlights the potential benefits of combining a short summary of the general goal with human descriptions, enabling the robot to make implicit inferences when possible. While such inferences can be advantageous in certain scenarios, they might not be suitable in others. We believe that refining our prompts could further reduce the error rate. For example, if the goal was personalization, we could emphasize the importance of the individual human description and ask the LLM to not draw implicit inferences based on the overarching objective. When accounting for both the global goal and the human descriptions, the error rate was further reduced to lower single digits. Interestingly, the LLM most frequently misinterpreted the 'speed' feature. This could be attributed to the inherent challenge of balancing two objectives: reaching the destination (faster speed) and ensuring perceived safety (slower speed). Following this, one natural limitation is that successful parsing of language descriptions depends on how well the prompt is formulated. Finally, our comparison uses 'preferences only' as the baseline.

\vspace{-5pt}
\section{Conclusion}
In this paper, we introduced PREDILECT, a framework that leverages the zero-shot capabilities of LLMs to expand the amount of information available, when inferring a reward function from human preferences, in an effort to mitigate causal confusion.
By combining preferences with language we can extract more information per query while offering a natural way of interacting with robots. We believe that the combination of these modalities is a promising approach to improve learning from human feedback.

%% file: content/08appendix.tex
\section{Appendix}
\subsection{Implementation details}

For our implementation, we use a PPO \cite{ schulman2017proximal} algorithm implementation from Stable Baselines \cite{stable-baselines}. We modified the normal RL loop to make it conform to our preference learning implementation as described in the main paper. We implemented the reward model from scratch based on details from Cristiano et al. \citep{christiano2017deep} in PyTorch\citep{NEURIPS2019_9015}. In addition, we extend the reward model to enable the usage of highlights as an additional learning objective.

To collect segments, we temporarily stop training and execute the current policy using undirected homogenous exploration \cite{wirth2017survey}. The segments are collected sequentially with a randomly sized interval to increase the diversity. We handle terminal states by restarting the environment while continuing the same segment. Another considered method was to end the segment and fill the rest with some value without meaning. After consideration, we chose the restart method to ensure that no information about the terminal state was leaked into the segments. We used uniform random sampling to generate the queries.

The reward function's architecture includes hidden layers with dimensions of (256, 256, 256). These layers primarily use ReLU activation functions with the output layer using a tanh activation. Both the agent's policy and value function have a structure of (128, 128) and use ReLU activations.

The reward function's network architecture was designed as a neural network. This network had hidden layers with dimensions of (255, 255, 255) and primarily employed ReLU activation functions. Only the final layer used a tanh activation, ensuring output values remained between -1 and 1. The agent's policy and value network was of size (128, 128) for Reacher and (256, 256) for Cheetah. Both were using ReLU activations for the hidden layers.

In both Cheetah and Reacher, every 20K timesteps, the agent updates its reward function. With each update, we use 10\% of the total available queries until it used up the query quota.

\subsection{Hyperparameters}
\label{sec:hyperparemeters}

\begin{table}[htb] 
\centering 

\caption{Reward hyperparameters}
\begin{tabular}{llllllll}
    \hline
    \multicolumn{1}{c}{\textbf{Value}}\\
    \hline
    \multicolumn{1}{c}{\textbf{Network architecture}}\\
    \hline
    \{hidden,output\} activation &  \{relu,tanh\} \\
    Hidden sizes & \{256,256,256\} \\
    
    \hline
    \multicolumn{1}{c}{\textbf{Preference Learning}}\\
    \hline
    Segment size & 50 \\
    Initial amount of queries & 1/10 of queries \\
    Initial training epochs & 200 \\
    Queries per update & 1/10 of queries \\
    Training epochs per update & 50 \\
    Batch size & 128 \\
    Timesteps between updates & 20K \\
    Learning rate & 0.0003 \\
    \hline

    \hline

  \end{tabular}
  \label{tab:hyperparameters}
\end{table}

\begin{table}[htb]
\centering 

\caption{PPO hyperparameters}
\begin{tabular}{llllllll}
    \hline
    \multicolumn{3}{c}{\textbf{Value}}\\
    & Walker2D & Cheetah & Social Nav.  \\
    \hline
    \multicolumn{4}{c}{\textbf{Network architecture}}\\
    \hline
    \{hidden,output\} act.(Actor) & \{tanh,tanh\}  & \{relu,tanh\} &  \{relu,tanh\}  \\
    \{hidden,output\} act.(Critic) & \{tanh,tanh\}  & \{relu,tanh\} &  \{relu,tanh\}  \\
    Hidden layers(Actor) & \{64,64\} & \{256,256\} & \{128,128\} \\
    Hidden layers(Critic) & \{64,64\} & \{256,256\} & \{128,128\} \\

    \hline
    \multicolumn{4}{c}{\textbf{RL parameters}}\\
    \hline
    Critic learning rate & 5e-05 & 2e-05 & 3e-04 \\
    Actor learning rate & 5e-05 & 2e-05 & 3e-04 \\
    Batch size & 32 & 64 & 128 \\
    Discount factor & 0.99 & 0.98 & 0.99 \\
    n steps & 512 & 512 & 1024 \\
    n epochs  & 20 & 20 & 10 \\
    GAE lambda ($\tau$) & 0.95 & 0.92 & 0.99 \\
    Clip range & 0.1 & 0.2 & 0.2 \\
    Normalized advantage & True & True & True\\
    Entropy coefficient & 6e-04 & 4e-04 & 5e-04 \\
    VF coefficient & 0.87 & 0.5 & 0.5 \\

    \hline

  \end{tabular}
  \label{tab:hyperparameters}
\end{table}

%% file: HRI2022.bbl

\begin{thebibliography}{80}


\ifx \showCODEN    \undefined \def \showCODEN     #1{\unskip}     \fi
\ifx \showDOI      \undefined \def \showDOI       #1{#1}\fi
\ifx \showISBNx    \undefined \def \showISBNx     #1{\unskip}     \fi
\ifx \showISBNxiii \undefined \def \showISBNxiii  #1{\unskip}     \fi
\ifx \showISSN     \undefined \def \showISSN      #1{\unskip}     \fi
\ifx \showLCCN     \undefined \def \showLCCN      #1{\unskip}     \fi
\ifx \shownote     \undefined \def \shownote      #1{#1}          \fi
\ifx \showarticletitle \undefined \def \showarticletitle #1{#1}   \fi
\ifx \showURL      \undefined \def \showURL       {\relax}        \fi
\providecommand\bibfield[2]{#2}
\providecommand\bibinfo[2]{#2}
\providecommand\natexlab[1]{#1}
\providecommand\showeprint[2][]{arXiv:#2}

\bibitem[Amin et~al\mbox{.}(2017)]%
        {amin2017repeated}
\bibfield{author}{\bibinfo{person}{Kareem Amin}, \bibinfo{person}{Nan Jiang}, {and} \bibinfo{person}{Satinder Singh}.} \bibinfo{year}{2017}\natexlab{}.
\newblock \showarticletitle{Repeated inverse reinforcement learning}.
\newblock \bibinfo{journal}{\emph{Advances in neural information processing systems}}  \bibinfo{volume}{30} (\bibinfo{year}{2017}).
\newblock


\bibitem[Amodei et~al\mbox{.}(2016)]%
        {amodei2016concrete}
\bibfield{author}{\bibinfo{person}{Dario Amodei}, \bibinfo{person}{Chris Olah}, \bibinfo{person}{Jacob Steinhardt}, \bibinfo{person}{Paul Christiano}, \bibinfo{person}{John Schulman}, {and} \bibinfo{person}{Dan Man{\'e}}.} \bibinfo{year}{2016}\natexlab{}.
\newblock \showarticletitle{Concrete problems in AI safety}.
\newblock \bibinfo{journal}{\emph{arXiv preprint arXiv:1606.06565}} (\bibinfo{year}{2016}).
\newblock


\bibitem[Basu et~al\mbox{.}(2018)]%
        {basu2018learning}
\bibfield{author}{\bibinfo{person}{Chandrayee Basu}, \bibinfo{person}{Mukesh Singhal}, {and} \bibinfo{person}{Anca~D Dragan}.} \bibinfo{year}{2018}\natexlab{}.
\newblock \showarticletitle{Learning from richer human guidance: Augmenting comparison-based learning with feature queries}. In \bibinfo{booktitle}{\emph{Proceedings of the 2018 ACM/IEEE International Conference on Human-Robot Interaction}}. \bibinfo{pages}{132--140}.
\newblock


\bibitem[B{\aa}venstrand and Berggren(2019)]%
        {baavenstrand2019performance}
\bibfield{author}{\bibinfo{person}{Erik B{\aa}venstrand} {and} \bibinfo{person}{Jakob Berggren}.} \bibinfo{year}{2019}\natexlab{}.
\newblock \bibinfo{title}{Performance evaluation of imitation learning algorithms with human experts}.
\newblock
\newblock


\bibitem[Bhatia et~al\mbox{.}(2020)]%
        {bhatia2020preference}
\bibfield{author}{\bibinfo{person}{Kush Bhatia}, \bibinfo{person}{Ashwin Pananjady}, \bibinfo{person}{Peter Bartlett}, \bibinfo{person}{Anca Dragan}, {and} \bibinfo{person}{Martin~J Wainwright}.} \bibinfo{year}{2020}\natexlab{}.
\newblock \showarticletitle{Preference learning along multiple criteria: A game-theoretic perspective}.
\newblock \bibinfo{journal}{\emph{Advances in neural information processing systems}}  \bibinfo{volume}{33} (\bibinfo{year}{2020}), \bibinfo{pages}{7413--7424}.
\newblock


\bibitem[B{\i}y{\i}k et~al\mbox{.}(2020)]%
        {biyik2020active}
\bibfield{author}{\bibinfo{person}{Erdem B{\i}y{\i}k}, \bibinfo{person}{Nicolas Huynh}, \bibinfo{person}{Mykel~J Kochenderfer}, {and} \bibinfo{person}{Dorsa Sadigh}.} \bibinfo{year}{2020}\natexlab{}.
\newblock \showarticletitle{Active preference-based gaussian process regression for reward learning}. In \bibinfo{booktitle}{\emph{Robotics: Science and Systems (RSS)}}.
\newblock


\bibitem[Bobu et~al\mbox{.}(2020)]%
        {bobu2020less}
\bibfield{author}{\bibinfo{person}{Andreea Bobu}, \bibinfo{person}{Dexter~RR Scobee}, \bibinfo{person}{Jaime~F Fisac}, \bibinfo{person}{S~Shankar Sastry}, {and} \bibinfo{person}{Anca~D Dragan}.} \bibinfo{year}{2020}\natexlab{}.
\newblock \showarticletitle{Less is more: Rethinking probabilistic models of human behavior}. In \bibinfo{booktitle}{\emph{Proceedings of the 2020 acm/ieee international conference on human-robot interaction}}. \bibinfo{pages}{429--437}.
\newblock


\bibitem[Bommasani et~al\mbox{.}(2021)]%
        {bommasani2021opportunities}
\bibfield{author}{\bibinfo{person}{Rishi Bommasani}, \bibinfo{person}{Drew~A Hudson}, \bibinfo{person}{Ehsan Adeli}, \bibinfo{person}{Russ Altman}, \bibinfo{person}{Simran Arora}, \bibinfo{person}{Sydney von Arx}, \bibinfo{person}{Michael~S Bernstein}, \bibinfo{person}{Jeannette Bohg}, \bibinfo{person}{Antoine Bosselut}, \bibinfo{person}{Emma Brunskill}, {et~al\mbox{.}}} \bibinfo{year}{2021}\natexlab{}.
\newblock \showarticletitle{On the opportunities and risks of foundation models}.
\newblock \bibinfo{journal}{\emph{arXiv preprint arXiv:2108.07258}} (\bibinfo{year}{2021}).
\newblock


\bibitem[Booth et~al\mbox{.}(2022)]%
        {booth2022revisiting}
\bibfield{author}{\bibinfo{person}{Serena Booth}, \bibinfo{person}{Sanjana Sharma}, \bibinfo{person}{Sarah Chung}, \bibinfo{person}{Julie Shah}, {and} \bibinfo{person}{Elena~L Glassman}.} \bibinfo{year}{2022}\natexlab{}.
\newblock \showarticletitle{Revisiting Human-Robot Teaching and Learning Through the Lens of Human Concept Learning}. In \bibinfo{booktitle}{\emph{Proceedings of the 2022 ACM/IEEE International Conference on Human-Robot Interaction}}. \bibinfo{pages}{147--156}.
\newblock


\bibitem[Bradley and Terry(1952)]%
        {bradley1952rank}
\bibfield{author}{\bibinfo{person}{Ralph~Allan Bradley} {and} \bibinfo{person}{Milton~E Terry}.} \bibinfo{year}{1952}\natexlab{}.
\newblock \showarticletitle{Rank analysis of incomplete block designs: I. The method of paired comparisons}.
\newblock \bibinfo{journal}{\emph{Biometrika}} \bibinfo{volume}{39}, \bibinfo{number}{3/4} (\bibinfo{year}{1952}), \bibinfo{pages}{324--345}.
\newblock


\bibitem[Brockman et~al\mbox{.}(2016)]%
        {brockman2016openai}
\bibfield{author}{\bibinfo{person}{Greg Brockman}, \bibinfo{person}{Vicki Cheung}, \bibinfo{person}{Ludwig Pettersson}, \bibinfo{person}{Jonas Schneider}, \bibinfo{person}{John Schulman}, \bibinfo{person}{Jie Tang}, {and} \bibinfo{person}{Wojciech Zaremba}.} \bibinfo{year}{2016}\natexlab{}.
\newblock \showarticletitle{Openai gym}.
\newblock \bibinfo{journal}{\emph{arXiv preprint arXiv:1606.01540}} (\bibinfo{year}{2016}).
\newblock


\bibitem[Brown et~al\mbox{.}(2020)]%
        {brown2020language}
\bibfield{author}{\bibinfo{person}{Tom Brown}, \bibinfo{person}{Benjamin Mann}, \bibinfo{person}{Nick Ryder}, \bibinfo{person}{Melanie Subbiah}, \bibinfo{person}{Jared~D Kaplan}, \bibinfo{person}{Prafulla Dhariwal}, \bibinfo{person}{Arvind Neelakantan}, \bibinfo{person}{Pranav Shyam}, \bibinfo{person}{Girish Sastry}, \bibinfo{person}{Amanda Askell}, {et~al\mbox{.}}} \bibinfo{year}{2020}\natexlab{}.
\newblock \showarticletitle{Language models are few-shot learners}.
\newblock \bibinfo{journal}{\emph{Advances in neural information processing systems}}  \bibinfo{volume}{33} (\bibinfo{year}{2020}), \bibinfo{pages}{1877--1901}.
\newblock


\bibitem[Cakmak et~al\mbox{.}(2011)]%
        {cakmak2011human}
\bibfield{author}{\bibinfo{person}{Maya Cakmak}, \bibinfo{person}{Siddhartha~S Srinivasa}, \bibinfo{person}{Min~Kyung Lee}, \bibinfo{person}{Jodi Forlizzi}, {and} \bibinfo{person}{Sara Kiesler}.} \bibinfo{year}{2011}\natexlab{}.
\newblock \showarticletitle{Human preferences for robot-human hand-over configurations}. In \bibinfo{booktitle}{\emph{2011 IEEE/RSJ International Conference on Intelligent Robots and Systems}}. IEEE, \bibinfo{pages}{1986--1993}.
\newblock


\bibitem[Chen et~al\mbox{.}(2021)]%
        {chen2021evaluating}
\bibfield{author}{\bibinfo{person}{Mark Chen}, \bibinfo{person}{Jerry Tworek}, \bibinfo{person}{Heewoo Jun}, \bibinfo{person}{Qiming Yuan}, \bibinfo{person}{Henrique Ponde de~Oliveira Pinto}, \bibinfo{person}{Jared Kaplan}, \bibinfo{person}{Harri Edwards}, \bibinfo{person}{Yuri Burda}, \bibinfo{person}{Nicholas Joseph}, \bibinfo{person}{Greg Brockman}, {et~al\mbox{.}}} \bibinfo{year}{2021}\natexlab{}.
\newblock \showarticletitle{Evaluating large language models trained on code}.
\newblock \bibinfo{journal}{\emph{arXiv preprint arXiv:2107.03374}} (\bibinfo{year}{2021}).
\newblock


\bibitem[Chen et~al\mbox{.}(2020)]%
        {chen2020semi}
\bibfield{author}{\bibinfo{person}{Yanbei Chen}, \bibinfo{person}{Xiatian Zhu}, \bibinfo{person}{Wei Li}, {and} \bibinfo{person}{Shaogang Gong}.} \bibinfo{year}{2020}\natexlab{}.
\newblock \showarticletitle{Semi-supervised learning under class distribution mismatch}. In \bibinfo{booktitle}{\emph{Proceedings of the AAAI Conference on Artificial Intelligence}}, Vol.~\bibinfo{volume}{34}. \bibinfo{pages}{3569--3576}.
\newblock


\bibitem[Christiano et~al\mbox{.}(2017)]%
        {christiano2017deep}
\bibfield{author}{\bibinfo{person}{Paul~F Christiano}, \bibinfo{person}{Jan Leike}, \bibinfo{person}{Tom Brown}, \bibinfo{person}{Miljan Martic}, \bibinfo{person}{Shane Legg}, {and} \bibinfo{person}{Dario Amodei}.} \bibinfo{year}{2017}\natexlab{}.
\newblock \showarticletitle{Deep reinforcement learning from human preferences}.
\newblock \bibinfo{journal}{\emph{Advances in neural information processing systems}}  \bibinfo{volume}{30} (\bibinfo{year}{2017}).
\newblock


\bibitem[Cui and Niekum(2018)]%
        {cui2018active}
\bibfield{author}{\bibinfo{person}{Yuchen Cui} {and} \bibinfo{person}{Scott Niekum}.} \bibinfo{year}{2018}\natexlab{}.
\newblock \showarticletitle{Active reward learning from critiques}. In \bibinfo{booktitle}{\emph{2018 IEEE international conference on robotics and automation (ICRA)}}. IEEE, \bibinfo{pages}{6907--6914}.
\newblock


\bibitem[De~Bruin et~al\mbox{.}(2018)]%
        {de2018integrating}
\bibfield{author}{\bibinfo{person}{Tim De~Bruin}, \bibinfo{person}{Jens Kober}, \bibinfo{person}{Karl Tuyls}, {and} \bibinfo{person}{Robert Babu{\v{s}}ka}.} \bibinfo{year}{2018}\natexlab{}.
\newblock \showarticletitle{Integrating state representation learning into deep reinforcement learning}.
\newblock \bibinfo{journal}{\emph{IEEE Robotics and Automation Letters}} \bibinfo{volume}{3}, \bibinfo{number}{3} (\bibinfo{year}{2018}), \bibinfo{pages}{1394--1401}.
\newblock


\bibitem[De~Haan et~al\mbox{.}(2019)]%
        {de2019causal}
\bibfield{author}{\bibinfo{person}{Pim De~Haan}, \bibinfo{person}{Dinesh Jayaraman}, {and} \bibinfo{person}{Sergey Levine}.} \bibinfo{year}{2019}\natexlab{}.
\newblock \showarticletitle{Causal confusion in imitation learning}.
\newblock \bibinfo{journal}{\emph{Advances in Neural Information Processing Systems}}  \bibinfo{volume}{32} (\bibinfo{year}{2019}).
\newblock


\bibitem[Devlin et~al\mbox{.}(2018)]%
        {devlin2018bert}
\bibfield{author}{\bibinfo{person}{Jacob Devlin}, \bibinfo{person}{Ming-Wei Chang}, \bibinfo{person}{Kenton Lee}, {and} \bibinfo{person}{Kristina Toutanova}.} \bibinfo{year}{2018}\natexlab{}.
\newblock \showarticletitle{Bert: Pre-training of deep bidirectional transformers for language understanding}.
\newblock \bibinfo{journal}{\emph{arXiv preprint arXiv:1810.04805}} (\bibinfo{year}{2018}).
\newblock


\bibitem[Eberhardt(2017)]%
        {eberhardt2017introduction}
\bibfield{author}{\bibinfo{person}{Frederick Eberhardt}.} \bibinfo{year}{2017}\natexlab{}.
\newblock \showarticletitle{Introduction to the foundations of causal discovery}.
\newblock \bibinfo{journal}{\emph{International Journal of Data Science and Analytics}} \bibinfo{volume}{3}, \bibinfo{number}{2} (\bibinfo{year}{2017}), \bibinfo{pages}{81--91}.
\newblock


\bibitem[Florence et~al\mbox{.}(2019)]%
        {florence2019self}
\bibfield{author}{\bibinfo{person}{Peter Florence}, \bibinfo{person}{Lucas Manuelli}, {and} \bibinfo{person}{Russ Tedrake}.} \bibinfo{year}{2019}\natexlab{}.
\newblock \showarticletitle{Self-supervised correspondence in visuomotor policy learning}.
\newblock \bibinfo{journal}{\emph{IEEE Robotics and Automation Letters}} \bibinfo{volume}{5}, \bibinfo{number}{2} (\bibinfo{year}{2019}), \bibinfo{pages}{492--499}.
\newblock


\bibitem[Francis et~al\mbox{.}(2023)]%
        {francis2023principles}
\bibfield{author}{\bibinfo{person}{Anthony Francis}, \bibinfo{person}{Claudia P{\'e}rez-d'Arpino}, \bibinfo{person}{Chengshu Li}, \bibinfo{person}{Fei Xia}, \bibinfo{person}{Alexandre Alahi}, \bibinfo{person}{Rachid Alami}, \bibinfo{person}{Aniket Bera}, \bibinfo{person}{Abhijat Biswas}, \bibinfo{person}{Joydeep Biswas}, \bibinfo{person}{Rohan Chandra}, {et~al\mbox{.}}} \bibinfo{year}{2023}\natexlab{}.
\newblock \showarticletitle{Principles and guidelines for evaluating social robot navigation algorithms}.
\newblock \bibinfo{journal}{\emph{arXiv preprint arXiv:2306.16740}} (\bibinfo{year}{2023}).
\newblock


\bibitem[Gavves et~al\mbox{.}(2015)]%
        {gavves2015active}
\bibfield{author}{\bibinfo{person}{Efstratios Gavves}, \bibinfo{person}{Thomas Mensink}, \bibinfo{person}{Tatiana Tommasi}, \bibinfo{person}{Cees~GM Snoek}, {and} \bibinfo{person}{Tinne Tuytelaars}.} \bibinfo{year}{2015}\natexlab{}.
\newblock \showarticletitle{Active transfer learning with zero-shot priors: Reusing past datasets for future tasks}. In \bibinfo{booktitle}{\emph{Proceedings of the IEEE International Conference on Computer Vision}}. \bibinfo{pages}{2731--2739}.
\newblock


\bibitem[Hadfield-Menell et~al\mbox{.}(2017)]%
        {hadfield2017inverse}
\bibfield{author}{\bibinfo{person}{Dylan Hadfield-Menell}, \bibinfo{person}{Smitha Milli}, \bibinfo{person}{Pieter Abbeel}, \bibinfo{person}{Stuart~J Russell}, {and} \bibinfo{person}{Anca Dragan}.} \bibinfo{year}{2017}\natexlab{}.
\newblock \showarticletitle{Inverse reward design}.
\newblock \bibinfo{journal}{\emph{Advances in neural information processing systems}}  \bibinfo{volume}{30} (\bibinfo{year}{2017}).
\newblock


\bibitem[Hejna~III and Sadigh(2023)]%
        {hejna2023few}
\bibfield{author}{\bibinfo{person}{Donald~Joseph Hejna~III} {and} \bibinfo{person}{Dorsa Sadigh}.} \bibinfo{year}{2023}\natexlab{}.
\newblock \showarticletitle{Few-shot preference learning for human-in-the-loop rl}. In \bibinfo{booktitle}{\emph{Conference on Robot Learning}}. PMLR, \bibinfo{pages}{2014--2025}.
\newblock


\bibitem[Helbing and Molnár(1995)]%
        {Helbing_1995}
\bibfield{author}{\bibinfo{person}{Dirk Helbing} {and} \bibinfo{person}{Péter Molnár}.} \bibinfo{year}{1995}\natexlab{}.
\newblock \showarticletitle{Social force model for pedestrian dynamics}.
\newblock \bibinfo{journal}{\emph{Physical Review E}} \bibinfo{volume}{51}, \bibinfo{number}{5} (\bibinfo{year}{1995}), \bibinfo{pages}{4282–4286}.
\newblock


\bibitem[Hill et~al\mbox{.}(2018)]%
        {stable-baselines}
\bibfield{author}{\bibinfo{person}{Ashley Hill}, \bibinfo{person}{Antonin Raffin}, \bibinfo{person}{Maximilian Ernestus}, \bibinfo{person}{Adam Gleave}, \bibinfo{person}{Anssi Kanervisto}, \bibinfo{person}{Rene Traore}, \bibinfo{person}{Prafulla Dhariwal}, \bibinfo{person}{Christopher Hesse}, \bibinfo{person}{Oleg Klimov}, \bibinfo{person}{Alex Nichol}, \bibinfo{person}{Matthias Plappert}, \bibinfo{person}{Alec Radford}, \bibinfo{person}{John Schulman}, \bibinfo{person}{Szymon Sidor}, {and} \bibinfo{person}{Yuhuai Wu}.} \bibinfo{year}{2018}\natexlab{}.
\newblock \bibinfo{title}{Stable Baselines}.
\newblock \bibinfo{howpublished}{\url{https://github.com/hill-a/stable-baselines}}.
\newblock


\bibitem[Ho et~al\mbox{.}(2022)]%
        {ho2022large}
\bibfield{author}{\bibinfo{person}{Namgyu Ho}, \bibinfo{person}{Laura Schmid}, {and} \bibinfo{person}{Se-Young Yun}.} \bibinfo{year}{2022}\natexlab{}.
\newblock \showarticletitle{Large language models are reasoning teachers}.
\newblock \bibinfo{journal}{\emph{arXiv preprint arXiv:2212.10071}} (\bibinfo{year}{2022}).
\newblock


\bibitem[Hu et~al\mbox{.}(2022)]%
        {hu2022explaining}
\bibfield{author}{\bibinfo{person}{Robert Hu}, \bibinfo{person}{Siu~Lun Chau}, \bibinfo{person}{Jaime~Ferrando Huertas}, {and} \bibinfo{person}{Dino Sejdinovic}.} \bibinfo{year}{2022}\natexlab{}.
\newblock \showarticletitle{Explaining Preferences with Shapley Values}. In \bibinfo{booktitle}{\emph{Advances in Neural Information Processing Systems}}, \bibfield{editor}{\bibinfo{person}{Alice~H. Oh}, \bibinfo{person}{Alekh Agarwal}, \bibinfo{person}{Danielle Belgrave}, {and} \bibinfo{person}{Kyunghyun Cho}} (Eds.).
\newblock
\urldef\tempurl%
\url{https://openreview.net/forum?id=-me36V0os8P}
\showURL{%
\tempurl}


\bibitem[Ibarz et~al\mbox{.}(2018)]%
        {ibarz2018reward}
\bibfield{author}{\bibinfo{person}{Borja Ibarz}, \bibinfo{person}{Jan Leike}, \bibinfo{person}{Tobias Pohlen}, \bibinfo{person}{Geoffrey Irving}, \bibinfo{person}{Shane Legg}, {and} \bibinfo{person}{Dario Amodei}.} \bibinfo{year}{2018}\natexlab{}.
\newblock \showarticletitle{Reward learning from human preferences and demonstrations in atari}.
\newblock \bibinfo{journal}{\emph{Advances in neural information processing systems}}  \bibinfo{volume}{31} (\bibinfo{year}{2018}).
\newblock


\bibitem[Jaderberg et~al\mbox{.}(2016)]%
        {jaderberg2016reinforcement}
\bibfield{author}{\bibinfo{person}{Max Jaderberg}, \bibinfo{person}{Volodymyr Mnih}, \bibinfo{person}{Wojciech~Marian Czarnecki}, \bibinfo{person}{Tom Schaul}, \bibinfo{person}{Joel~Z Leibo}, \bibinfo{person}{David Silver}, {and} \bibinfo{person}{Koray Kavukcuoglu}.} \bibinfo{year}{2016}\natexlab{}.
\newblock \showarticletitle{Reinforcement learning with unsupervised auxiliary tasks}.
\newblock \bibinfo{journal}{\emph{arXiv preprint arXiv:1611.05397}} (\bibinfo{year}{2016}).
\newblock


\bibitem[Jeon et~al\mbox{.}(2020)]%
        {jeon2020reward}
\bibfield{author}{\bibinfo{person}{Hong~Jun Jeon}, \bibinfo{person}{Smitha Milli}, {and} \bibinfo{person}{Anca Dragan}.} \bibinfo{year}{2020}\natexlab{}.
\newblock \showarticletitle{Reward-rational (implicit) choice: A unifying formalism for reward learning}.
\newblock \bibinfo{journal}{\emph{Advances in Neural Information Processing Systems}}  \bibinfo{volume}{33} (\bibinfo{year}{2020}), \bibinfo{pages}{4415--4426}.
\newblock


\bibitem[Khayrallah et~al\mbox{.}(2015)]%
        {khayrallah2015natural}
\bibfield{author}{\bibinfo{person}{Huda Khayrallah}, \bibinfo{person}{Sean Trott}, {and} \bibinfo{person}{Jerome Feldman}.} \bibinfo{year}{2015}\natexlab{}.
\newblock \showarticletitle{Natural language for human robot interaction}. In \bibinfo{booktitle}{\emph{International Conference on Human-Robot Interaction (HRI)}}.
\newblock


\bibitem[Knox and Stone(2009)]%
        {knox2009interactively}
\bibfield{author}{\bibinfo{person}{W~Bradley Knox} {and} \bibinfo{person}{Peter Stone}.} \bibinfo{year}{2009}\natexlab{}.
\newblock \showarticletitle{Interactively shaping agents via human reinforcement: The TAMER framework}. In \bibinfo{booktitle}{\emph{Proceedings of the fifth international conference on Knowledge capture}}. \bibinfo{pages}{9--16}.
\newblock


\bibitem[Knox et~al\mbox{.}(2013)]%
        {knox2013training}
\bibfield{author}{\bibinfo{person}{W~Bradley Knox}, \bibinfo{person}{Peter Stone}, {and} \bibinfo{person}{Cynthia Breazeal}.} \bibinfo{year}{2013}\natexlab{}.
\newblock \showarticletitle{Training a robot via human feedback: A case study}. In \bibinfo{booktitle}{\emph{International Conference on Social Robotics}}. Springer, \bibinfo{pages}{460--470}.
\newblock


\bibitem[Kulkarni et~al\mbox{.}(2019)]%
        {kulkarni2019unsupervised}
\bibfield{author}{\bibinfo{person}{Tejas~D Kulkarni}, \bibinfo{person}{Ankush Gupta}, \bibinfo{person}{Catalin Ionescu}, \bibinfo{person}{Sebastian Borgeaud}, \bibinfo{person}{Malcolm Reynolds}, \bibinfo{person}{Andrew Zisserman}, {and} \bibinfo{person}{Volodymyr Mnih}.} \bibinfo{year}{2019}\natexlab{}.
\newblock \showarticletitle{Unsupervised learning of object keypoints for perception and control}.
\newblock \bibinfo{journal}{\emph{Advances in neural information processing systems}}  \bibinfo{volume}{32} (\bibinfo{year}{2019}).
\newblock


\bibitem[Kumar et~al\mbox{.}(2022)]%
        {kumar2022using}
\bibfield{author}{\bibinfo{person}{Sreejan Kumar}, \bibinfo{person}{Carlos~G Correa}, \bibinfo{person}{Ishita Dasgupta}, \bibinfo{person}{Raja Marjieh}, \bibinfo{person}{Michael~Y Hu}, \bibinfo{person}{Robert Hawkins}, \bibinfo{person}{Jonathan~D Cohen}, \bibinfo{person}{Karthik Narasimhan}, \bibinfo{person}{Tom Griffiths}, {et~al\mbox{.}}} \bibinfo{year}{2022}\natexlab{}.
\newblock \showarticletitle{Using natural language and program abstractions to instill human inductive biases in machines}.
\newblock \bibinfo{journal}{\emph{Advances in Neural Information Processing Systems}}  \bibinfo{volume}{35} (\bibinfo{year}{2022}), \bibinfo{pages}{167--180}.
\newblock


\bibitem[Lee et~al\mbox{.}(2021)]%
        {lee2021pebble}
\bibfield{author}{\bibinfo{person}{Kimin Lee}, \bibinfo{person}{Laura Smith}, {and} \bibinfo{person}{Pieter Abbeel}.} \bibinfo{year}{2021}\natexlab{}.
\newblock \showarticletitle{Pebble: Feedback-efficient interactive reinforcement learning via relabeling experience and unsupervised pre-training}.
\newblock \bibinfo{journal}{\emph{arXiv preprint arXiv:2106.05091}} (\bibinfo{year}{2021}).
\newblock


\bibitem[Li et~al\mbox{.}(2021)]%
        {li2021intention}
\bibfield{author}{\bibinfo{person}{Zhihao Li}, \bibinfo{person}{Yishan Mu}, \bibinfo{person}{Zhenglong Sun}, \bibinfo{person}{Sifan Song}, \bibinfo{person}{Jionglong Su}, {and} \bibinfo{person}{Jiaming Zhang}.} \bibinfo{year}{2021}\natexlab{}.
\newblock \showarticletitle{Intention understanding in human--robot interaction based on visual-NLP semantics}.
\newblock \bibinfo{journal}{\emph{Frontiers in Neurorobotics}}  \bibinfo{volume}{14} (\bibinfo{year}{2021}), \bibinfo{pages}{610139}.
\newblock


\bibitem[Liang et~al\mbox{.}(2022)]%
        {liang2022reward}
\bibfield{author}{\bibinfo{person}{Xinran Liang}, \bibinfo{person}{Katherine Shu}, \bibinfo{person}{Kimin Lee}, {and} \bibinfo{person}{Pieter Abbeel}.} \bibinfo{year}{2022}\natexlab{}.
\newblock \showarticletitle{Reward Uncertainty for Exploration in Preference-based Reinforcement Learning}. In \bibinfo{booktitle}{\emph{International Conference on Learning Representations}}.
\newblock
\urldef\tempurl%
\url{https://openreview.net/forum?id=OWZVD-l-ZrC}
\showURL{%
\tempurl}


\bibitem[Liu et~al\mbox{.}(2022)]%
        {liu2022metarewardnet}
\bibfield{author}{\bibinfo{person}{Runze Liu}, \bibinfo{person}{Fengshuo Bai}, \bibinfo{person}{Yali Du}, {and} \bibinfo{person}{Yaodong Yang}.} \bibinfo{year}{2022}\natexlab{}.
\newblock \showarticletitle{Meta-Reward-Net: Implicitly Differentiable Reward Learning for Preference-based Reinforcement Learning}. In \bibinfo{booktitle}{\emph{Advances in Neural Information Processing Systems}}, \bibfield{editor}{\bibinfo{person}{Alice~H. Oh}, \bibinfo{person}{Alekh Agarwal}, \bibinfo{person}{Danielle Belgrave}, {and} \bibinfo{person}{Kyunghyun Cho}} (Eds.).
\newblock
\urldef\tempurl%
\url{https://openreview.net/forum?id=OZKBReUF-wX}
\showURL{%
\tempurl}


\bibitem[MacGlashan et~al\mbox{.}(2017)]%
        {macglashan2017interactive}
\bibfield{author}{\bibinfo{person}{James MacGlashan}, \bibinfo{person}{Mark~K Ho}, \bibinfo{person}{Robert Loftin}, \bibinfo{person}{Bei Peng}, \bibinfo{person}{Guan Wang}, \bibinfo{person}{David~L Roberts}, \bibinfo{person}{Matthew~E Taylor}, {and} \bibinfo{person}{Michael~L Littman}.} \bibinfo{year}{2017}\natexlab{}.
\newblock \showarticletitle{Interactive learning from policy-dependent human feedback}. In \bibinfo{booktitle}{\emph{Int. Conf. on Machine Learning}}. PMLR, \bibinfo{pages}{2285--2294}.
\newblock


\bibitem[Matas et~al\mbox{.}(2018)]%
        {matas2018sim}
\bibfield{author}{\bibinfo{person}{Jan Matas}, \bibinfo{person}{Stephen James}, {and} \bibinfo{person}{Andrew~J Davison}.} \bibinfo{year}{2018}\natexlab{}.
\newblock \showarticletitle{Sim-to-real reinforcement learning for deformable object manipulation}. In \bibinfo{booktitle}{\emph{Conference on Robot Learning}}. PMLR, \bibinfo{pages}{734--743}.
\newblock


\bibitem[Mirowski et~al\mbox{.}(2016)]%
        {mirowski2016learning}
\bibfield{author}{\bibinfo{person}{Piotr Mirowski}, \bibinfo{person}{Razvan Pascanu}, \bibinfo{person}{Fabio Viola}, \bibinfo{person}{Hubert Soyer}, \bibinfo{person}{Andrew~J Ballard}, \bibinfo{person}{Andrea Banino}, \bibinfo{person}{Misha Denil}, \bibinfo{person}{Ross Goroshin}, \bibinfo{person}{Laurent Sifre}, \bibinfo{person}{Koray Kavukcuoglu}, {et~al\mbox{.}}} \bibinfo{year}{2016}\natexlab{}.
\newblock \showarticletitle{Learning to navigate in complex environments}.
\newblock \bibinfo{journal}{\emph{arXiv preprint arXiv:1611.03673}} (\bibinfo{year}{2016}).
\newblock


\bibitem[Mokady et~al\mbox{.}(2021)]%
        {mokady2021clipcap}
\bibfield{author}{\bibinfo{person}{Ron Mokady}, \bibinfo{person}{Amir Hertz}, {and} \bibinfo{person}{Amit~H Bermano}.} \bibinfo{year}{2021}\natexlab{}.
\newblock \showarticletitle{Clipcap: Clip prefix for image captioning}.
\newblock \bibinfo{journal}{\emph{arXiv preprint arXiv:2111.09734}} (\bibinfo{year}{2021}).
\newblock


\bibitem[Najar and Chetouani(2021)]%
        {najar2021reinforcement}
\bibfield{author}{\bibinfo{person}{Anis Najar} {and} \bibinfo{person}{Mohamed Chetouani}.} \bibinfo{year}{2021}\natexlab{}.
\newblock \showarticletitle{Reinforcement learning with human advice: a survey}.
\newblock \bibinfo{journal}{\emph{Frontiers in Robotics and AI}}  \bibinfo{volume}{8} (\bibinfo{year}{2021}), \bibinfo{pages}{584075}.
\newblock


\bibitem[Ngiam et~al\mbox{.}(2011)]%
        {ngiam2011multimodal}
\bibfield{author}{\bibinfo{person}{Jiquan Ngiam}, \bibinfo{person}{Aditya Khosla}, \bibinfo{person}{Mingyu Kim}, \bibinfo{person}{Juhan Nam}, \bibinfo{person}{Honglak Lee}, {and} \bibinfo{person}{Andrew~Y Ng}.} \bibinfo{year}{2011}\natexlab{}.
\newblock \showarticletitle{Multimodal deep learning}. In \bibinfo{booktitle}{\emph{ICML}}.
\newblock


\bibitem[Park et~al\mbox{.}(2022)]%
        {park2022surf}
\bibfield{author}{\bibinfo{person}{Jongjin Park}, \bibinfo{person}{Younggyo Seo}, \bibinfo{person}{Jinwoo Shin}, \bibinfo{person}{Honglak Lee}, \bibinfo{person}{Pieter Abbeel}, {and} \bibinfo{person}{Kimin Lee}.} \bibinfo{year}{2022}\natexlab{}.
\newblock \showarticletitle{{SURF}: Semi-supervised Reward Learning with Data Augmentation for Feedback-efficient Preference-based Reinforcement Learning}. In \bibinfo{booktitle}{\emph{International Conference on Learning Representations}}.
\newblock
\urldef\tempurl%
\url{https://openreview.net/forum?id=TfhfZLQ2EJO}
\showURL{%
\tempurl}


\bibitem[Paszke et~al\mbox{.}(2019)]%
        {NEURIPS2019_9015}
\bibfield{author}{\bibinfo{person}{Adam Paszke}, \bibinfo{person}{Sam Gross}, \bibinfo{person}{Francisco Massa}, \bibinfo{person}{Adam Lerer}, \bibinfo{person}{James Bradbury}, \bibinfo{person}{Gregory Chanan}, \bibinfo{person}{Trevor Killeen}, \bibinfo{person}{Zeming Lin}, \bibinfo{person}{Natalia Gimelshein}, \bibinfo{person}{Luca Antiga}, \bibinfo{person}{Alban Desmaison}, \bibinfo{person}{Andreas Kopf}, \bibinfo{person}{Edward Yang}, \bibinfo{person}{Zachary DeVito}, \bibinfo{person}{Martin Raison}, \bibinfo{person}{Alykhan Tejani}, \bibinfo{person}{Sasank Chilamkurthy}, \bibinfo{person}{Benoit Steiner}, \bibinfo{person}{Lu Fang}, \bibinfo{person}{Junjie Bai}, {and} \bibinfo{person}{Soumith Chintala}.} \bibinfo{year}{2019}\natexlab{}.
\newblock \showarticletitle{PyTorch: An Imperative Style, High-Performance Deep Learning Library}.
\newblock In \bibinfo{booktitle}{\emph{Advances in Neural Information Processing Systems 32}}, \bibfield{editor}{\bibinfo{person}{H.~Wallach}, \bibinfo{person}{H.~Larochelle}, \bibinfo{person}{A.~Beygelzimer}, \bibinfo{person}{F.~d\textquotesingle Alch\'{e}-Buc}, \bibinfo{person}{E.~Fox}, {and} \bibinfo{person}{R.~Garnett}} (Eds.). \bibinfo{publisher}{Curran Associates, Inc.}, \bibinfo{pages}{8024--8035}.
\newblock


\bibitem[Pate et~al\mbox{.}(2021)]%
        {pate2021natural}
\bibfield{author}{\bibinfo{person}{Seth Pate}, \bibinfo{person}{Wei Xu}, \bibinfo{person}{Ziyi Yang}, \bibinfo{person}{Maxwell Love}, \bibinfo{person}{Siddarth Ganguri}, {and} \bibinfo{person}{Lawson~LS Wong}.} \bibinfo{year}{2021}\natexlab{}.
\newblock \showarticletitle{Natural language for human-robot collaboration: Problems beyond language grounding}.
\newblock \bibinfo{journal}{\emph{arXiv preprint arXiv:2110.04441}} (\bibinfo{year}{2021}).
\newblock


\bibitem[Pinto and Gupta(2017)]%
        {pinto2017learning}
\bibfield{author}{\bibinfo{person}{Lerrel Pinto} {and} \bibinfo{person}{Abhinav Gupta}.} \bibinfo{year}{2017}\natexlab{}.
\newblock \showarticletitle{Learning to push by grasping: Using multiple tasks for effective learning}. In \bibinfo{booktitle}{\emph{2017 IEEE international conference on robotics and automation (ICRA)}}. IEEE, \bibinfo{pages}{2161--2168}.
\newblock


\bibitem[Radford et~al\mbox{.}(2021)]%
        {radford2021learning}
\bibfield{author}{\bibinfo{person}{Alec Radford}, \bibinfo{person}{Jong~Wook Kim}, \bibinfo{person}{Chris Hallacy}, \bibinfo{person}{Aditya Ramesh}, \bibinfo{person}{Gabriel Goh}, \bibinfo{person}{Sandhini Agarwal}, \bibinfo{person}{Girish Sastry}, \bibinfo{person}{Amanda Askell}, \bibinfo{person}{Pamela Mishkin}, \bibinfo{person}{Jack Clark}, {et~al\mbox{.}}} \bibinfo{year}{2021}\natexlab{}.
\newblock \showarticletitle{Learning transferable visual models from natural language supervision}. In \bibinfo{booktitle}{\emph{International Conference on Machine Learning}}. PMLR, \bibinfo{pages}{8748--8763}.
\newblock


\bibitem[Radosavovic et~al\mbox{.}(2018)]%
        {radosavovic2018data}
\bibfield{author}{\bibinfo{person}{Ilija Radosavovic}, \bibinfo{person}{Piotr Doll{\'a}r}, \bibinfo{person}{Ross Girshick}, \bibinfo{person}{Georgia Gkioxari}, {and} \bibinfo{person}{Kaiming He}.} \bibinfo{year}{2018}\natexlab{}.
\newblock \showarticletitle{Data distillation: Towards omni-supervised learning}. In \bibinfo{booktitle}{\emph{Proceedings of the IEEE conference on computer vision and pattern recognition}}. \bibinfo{pages}{4119--4128}.
\newblock


\bibitem[Rajpurkar et~al\mbox{.}(2018)]%
        {rajpurkar2018know}
\bibfield{author}{\bibinfo{person}{Pranav Rajpurkar}, \bibinfo{person}{Robin Jia}, {and} \bibinfo{person}{Percy Liang}.} \bibinfo{year}{2018}\natexlab{}.
\newblock \showarticletitle{Know what you don't know: Unanswerable questions for SQuAD}.
\newblock \bibinfo{journal}{\emph{arXiv preprint arXiv:1806.03822}} (\bibinfo{year}{2018}).
\newblock


\bibitem[Sadigh et~al\mbox{.}(2017)]%
        {sadigh2017active}
\bibfield{author}{\bibinfo{person}{Dorsa Sadigh}, \bibinfo{person}{Anca~D Dragan}, \bibinfo{person}{Shankar Sastry}, {and} \bibinfo{person}{Sanjit~A Seshia}.} \bibinfo{year}{2017}\natexlab{}.
\newblock \bibinfo{booktitle}{\emph{Active preference-based learning of reward functions}}.
\newblock


\bibitem[Schrum et~al\mbox{.}(2022)]%
        {schrum2022mind}
\bibfield{author}{\bibinfo{person}{Mariah~L Schrum}, \bibinfo{person}{Erin Hedlund-Botti}, \bibinfo{person}{Nina Moorman}, {and} \bibinfo{person}{Matthew~C Gombolay}.} \bibinfo{year}{2022}\natexlab{}.
\newblock \showarticletitle{MIND MELD: Personalized Meta-Learning for Robot-Centric Imitation Learning.}. In \bibinfo{booktitle}{\emph{HRI}}. \bibinfo{pages}{157--165}.
\newblock


\bibitem[Schulman et~al\mbox{.}(2017)]%
        {schulman2017proximal}
\bibfield{author}{\bibinfo{person}{John Schulman}, \bibinfo{person}{Filip Wolski}, \bibinfo{person}{Prafulla Dhariwal}, \bibinfo{person}{Alec Radford}, {and} \bibinfo{person}{Oleg Klimov}.} \bibinfo{year}{2017}\natexlab{}.
\newblock \showarticletitle{Proximal policy optimization algorithms}.
\newblock \bibinfo{journal}{\emph{arXiv preprint arXiv:1707.06347}} (\bibinfo{year}{2017}).
\newblock


\bibitem[Schwab et~al\mbox{.}(2018)]%
        {schwab2018zero}
\bibfield{author}{\bibinfo{person}{Devin Schwab}, \bibinfo{person}{Yifeng Zhu}, {and} \bibinfo{person}{Manuela Veloso}.} \bibinfo{year}{2018}\natexlab{}.
\newblock \showarticletitle{Zero shot transfer learning for robot soccer}. In \bibinfo{booktitle}{\emph{Proceedings of the 17th International Conference on Autonomous Agents and MultiAgent Systems}}. \bibinfo{pages}{2070--2072}.
\newblock


\bibitem[Senft et~al\mbox{.}(2017)]%
        {senft2017supervised}
\bibfield{author}{\bibinfo{person}{Emmanuel Senft}, \bibinfo{person}{Paul Baxter}, \bibinfo{person}{James Kennedy}, \bibinfo{person}{S{\'e}verin Lemaignan}, {and} \bibinfo{person}{Tony Belpaeme}.} \bibinfo{year}{2017}\natexlab{}.
\newblock \showarticletitle{Supervised autonomy for online learning in human-robot interaction}.
\newblock \bibinfo{journal}{\emph{Pattern Recognition Letters}}  \bibinfo{volume}{99} (\bibinfo{year}{2017}), \bibinfo{pages}{77--86}.
\newblock


\bibitem[Shah et~al\mbox{.}(2022)]%
        {shah2022offline}
\bibfield{author}{\bibinfo{person}{Dhruv Shah}, \bibinfo{person}{Arjun Bhorkar}, \bibinfo{person}{Hrishit Leen}, \bibinfo{person}{Ilya Kostrikov}, \bibinfo{person}{Nicholas Rhinehart}, {and} \bibinfo{person}{Sergey Levine}.} \bibinfo{year}{2022}\natexlab{}.
\newblock \showarticletitle{Offline Reinforcement Learning for Visual Navigation}. In \bibinfo{booktitle}{\emph{6th Annual Conference on Robot Learning}}.
\newblock
\urldef\tempurl%
\url{https://openreview.net/forum?id=uhIfIEIiWm_}
\showURL{%
\tempurl}


\bibitem[Sharma et~al\mbox{.}(2022)]%
        {sharma2022correcting}
\bibfield{author}{\bibinfo{person}{Pratyusha Sharma}, \bibinfo{person}{Balakumar Sundaralingam}, \bibinfo{person}{Valts Blukis}, \bibinfo{person}{Chris Paxton}, \bibinfo{person}{Tucker Hermans}, \bibinfo{person}{Antonio Torralba}, \bibinfo{person}{Jacob Andreas}, {and} \bibinfo{person}{Dieter Fox}.} \bibinfo{year}{2022}\natexlab{}.
\newblock \showarticletitle{Correcting robot plans with natural language feedback}. In \bibinfo{booktitle}{\emph{Robotics: Science and Systems (RSS)}}.
\newblock


\bibitem[Shelhamer et~al\mbox{.}(2016)]%
        {shelhamer2016loss}
\bibfield{author}{\bibinfo{person}{Evan Shelhamer}, \bibinfo{person}{Parsa Mahmoudieh}, \bibinfo{person}{Max Argus}, {and} \bibinfo{person}{Trevor Darrell}.} \bibinfo{year}{2016}\natexlab{}.
\newblock \showarticletitle{Loss is its own reward: Self-supervision for reinforcement learning}.
\newblock \bibinfo{journal}{\emph{arXiv preprint arXiv:1612.07307}} (\bibinfo{year}{2016}).
\newblock


\bibitem[Soh et~al\mbox{.}(2020)]%
        {soh2020meta}
\bibfield{author}{\bibinfo{person}{Jae~Woong Soh}, \bibinfo{person}{Sunwoo Cho}, {and} \bibinfo{person}{Nam~Ik Cho}.} \bibinfo{year}{2020}\natexlab{}.
\newblock \showarticletitle{Meta-transfer learning for zero-shot super-resolution}. In \bibinfo{booktitle}{\emph{Proceedings of the IEEE/CVF Conference on Computer Vision and Pattern Recognition}}. \bibinfo{pages}{3516--3525}.
\newblock


\bibitem[Spirtes et~al\mbox{.}(2000)]%
        {spirtes2000causation}
\bibfield{author}{\bibinfo{person}{Peter Spirtes}, \bibinfo{person}{Clark~N Glymour}, \bibinfo{person}{Richard Scheines}, {and} \bibinfo{person}{David Heckerman}.} \bibinfo{year}{2000}\natexlab{}.
\newblock \bibinfo{booktitle}{\emph{Causation, prediction, and search}}.
\newblock \bibinfo{publisher}{MIT press}.
\newblock


\bibitem[Sumers et~al\mbox{.}(2022)]%
        {sumers2022how}
\bibfield{author}{\bibinfo{person}{Theodore Sumers}, \bibinfo{person}{Robert~D. Hawkins}, \bibinfo{person}{Mark~K Ho}, \bibinfo{person}{Thomas~L. Griffiths}, {and} \bibinfo{person}{Dylan Hadfield-Menell}.} \bibinfo{year}{2022}\natexlab{}.
\newblock \showarticletitle{How to talk so {AI} will learn: Instructions, descriptions, and autonomy}. In \bibinfo{booktitle}{\emph{Advances in Neural Information Processing Systems}}, \bibfield{editor}{\bibinfo{person}{Alice~H. Oh}, \bibinfo{person}{Alekh Agarwal}, \bibinfo{person}{Danielle Belgrave}, {and} \bibinfo{person}{Kyunghyun Cho}} (Eds.).
\newblock
\urldef\tempurl%
\url{https://openreview.net/forum?id=ZLsZmNe1RDb}
\showURL{%
\tempurl}


\bibitem[Thoppilan et~al\mbox{.}(2022)]%
        {thoppilan2022lamda}
\bibfield{author}{\bibinfo{person}{Romal Thoppilan}, \bibinfo{person}{Daniel De~Freitas}, \bibinfo{person}{Jamie Hall}, \bibinfo{person}{Noam Shazeer}, \bibinfo{person}{Apoorv Kulshreshtha}, \bibinfo{person}{Heng-Tze Cheng}, \bibinfo{person}{Alicia Jin}, \bibinfo{person}{Taylor Bos}, \bibinfo{person}{Leslie Baker}, \bibinfo{person}{Yu Du}, {et~al\mbox{.}}} \bibinfo{year}{2022}\natexlab{}.
\newblock \showarticletitle{Lamda: Language models for dialog applications}.
\newblock \bibinfo{journal}{\emph{arXiv preprint arXiv:2201.08239}} (\bibinfo{year}{2022}).
\newblock


\bibitem[Tien et~al\mbox{.}(2022)]%
        {tien2022study}
\bibfield{author}{\bibinfo{person}{Jeremy Tien}, \bibinfo{person}{Jerry Zhi-Yang He}, \bibinfo{person}{Zackory Erickson}, \bibinfo{person}{Anca~D Dragan}, {and} \bibinfo{person}{Daniel Brown}.} \bibinfo{year}{2022}\natexlab{}.
\newblock \showarticletitle{A Study of Causal Confusion in Preference-Based Reward Learning}.
\newblock \bibinfo{journal}{\emph{arXiv preprint arXiv:2204.06601}} (\bibinfo{year}{2022}).
\newblock


\bibitem[Tsimpoukelli et~al\mbox{.}(2021)]%
        {tsimpoukelli2021multimodal}
\bibfield{author}{\bibinfo{person}{Maria Tsimpoukelli}, \bibinfo{person}{Jacob~L Menick}, \bibinfo{person}{Serkan Cabi}, \bibinfo{person}{SM Eslami}, \bibinfo{person}{Oriol Vinyals}, {and} \bibinfo{person}{Felix Hill}.} \bibinfo{year}{2021}\natexlab{}.
\newblock \showarticletitle{Multimodal few-shot learning with frozen language models}.
\newblock \bibinfo{journal}{\emph{Advances in Neural Information Processing Systems}}  \bibinfo{volume}{34} (\bibinfo{year}{2021}), \bibinfo{pages}{200--212}.
\newblock


\bibitem[Wang et~al\mbox{.}(2022)]%
        {wang2022skill}
\bibfield{author}{\bibinfo{person}{Xiaofei Wang}, \bibinfo{person}{Kimin Lee}, \bibinfo{person}{Kourosh Hakhamaneshi}, \bibinfo{person}{Pieter Abbeel}, {and} \bibinfo{person}{Michael Laskin}.} \bibinfo{year}{2022}\natexlab{}.
\newblock \showarticletitle{Skill preferences: Learning to extract and execute robotic skills from human feedback}. In \bibinfo{booktitle}{\emph{Conference on Robot Learning}}. PMLR, \bibinfo{pages}{1259--1268}.
\newblock


\bibitem[Wei et~al\mbox{.}(2022)]%
        {wei2022chain}
\bibfield{author}{\bibinfo{person}{Jason Wei}, \bibinfo{person}{Xuezhi Wang}, \bibinfo{person}{Dale Schuurmans}, \bibinfo{person}{Maarten Bosma}, \bibinfo{person}{Fei Xia}, \bibinfo{person}{Ed Chi}, \bibinfo{person}{Quoc~V Le}, \bibinfo{person}{Denny Zhou}, {et~al\mbox{.}}} \bibinfo{year}{2022}\natexlab{}.
\newblock \showarticletitle{Chain-of-thought prompting elicits reasoning in large language models}.
\newblock \bibinfo{journal}{\emph{Advances in Neural Information Processing Systems}}  \bibinfo{volume}{35} (\bibinfo{year}{2022}), \bibinfo{pages}{24824--24837}.
\newblock


\bibitem[Wilson et~al\mbox{.}(2012)]%
        {wilson2012bayesian}
\bibfield{author}{\bibinfo{person}{Aaron Wilson}, \bibinfo{person}{Alan Fern}, {and} \bibinfo{person}{Prasad Tadepalli}.} \bibinfo{year}{2012}\natexlab{}.
\newblock \showarticletitle{A bayesian approach for policy learning from trajectory preference queries}.
\newblock \bibinfo{journal}{\emph{Advances in neural information processing systems}}  \bibinfo{volume}{25} (\bibinfo{year}{2012}).
\newblock


\bibitem[Wirth et~al\mbox{.}(2017)]%
        {wirth2017survey}
\bibfield{author}{\bibinfo{person}{Christian Wirth}, \bibinfo{person}{Riad Akrour}, \bibinfo{person}{Gerhard Neumann}, \bibinfo{person}{Johannes F{\"u}rnkranz}, {et~al\mbox{.}}} \bibinfo{year}{2017}\natexlab{}.
\newblock \showarticletitle{A survey of preference-based reinforcement learning methods}.
\newblock \bibinfo{journal}{\emph{Journal of Machine Learning Research}} \bibinfo{volume}{18}, \bibinfo{number}{136} (\bibinfo{year}{2017}), \bibinfo{pages}{1--46}.
\newblock


\bibitem[Xie et~al\mbox{.}(2018)]%
        {xie2018few}
\bibfield{author}{\bibinfo{person}{Annie Xie}, \bibinfo{person}{Avi Singh}, \bibinfo{person}{Sergey Levine}, {and} \bibinfo{person}{Chelsea Finn}.} \bibinfo{year}{2018}\natexlab{}.
\newblock \showarticletitle{Few-shot goal inference for visuomotor learning and planning}. In \bibinfo{booktitle}{\emph{Conference on Robot Learning}}. PMLR, \bibinfo{pages}{40--52}.
\newblock


\bibitem[Xie et~al\mbox{.}(2020)]%
        {xie2020self}
\bibfield{author}{\bibinfo{person}{Qizhe Xie}, \bibinfo{person}{Minh-Thang Luong}, \bibinfo{person}{Eduard Hovy}, {and} \bibinfo{person}{Quoc~V Le}.} \bibinfo{year}{2020}\natexlab{}.
\newblock \showarticletitle{Self-training with noisy student improves imagenet classification}. In \bibinfo{booktitle}{\emph{Proceedings of the IEEE/CVF conference on computer vision and pattern recognition}}. \bibinfo{pages}{10687--10698}.
\newblock


\bibitem[Ying et~al\mbox{.}(2018)]%
        {ying2018transfer}
\bibfield{author}{\bibinfo{person}{Wei Ying}, \bibinfo{person}{Yu Zhang}, \bibinfo{person}{Junzhou Huang}, {and} \bibinfo{person}{Qiang Yang}.} \bibinfo{year}{2018}\natexlab{}.
\newblock \showarticletitle{Transfer learning via learning to transfer}. In \bibinfo{booktitle}{\emph{International Conference on Machine Learning}}. PMLR, \bibinfo{pages}{5085--5094}.
\newblock


\bibitem[Zakka et~al\mbox{.}(2022)]%
        {zakka2022xirl}
\bibfield{author}{\bibinfo{person}{Kevin Zakka}, \bibinfo{person}{Andy Zeng}, \bibinfo{person}{Pete Florence}, \bibinfo{person}{Jonathan Tompson}, \bibinfo{person}{Jeannette Bohg}, {and} \bibinfo{person}{Debidatta Dwibedi}.} \bibinfo{year}{2022}\natexlab{}.
\newblock \showarticletitle{Xirl: Cross-embodiment inverse reinforcement learning}. In \bibinfo{booktitle}{\emph{Conference on Robot Learning}}. PMLR, \bibinfo{pages}{537--546}.
\newblock


\bibitem[Zeng et~al\mbox{.}(2022)]%
        {zeng2022socratic}
\bibfield{author}{\bibinfo{person}{Andy Zeng}, \bibinfo{person}{Maria Attarian}, \bibinfo{person}{Krzysztof~Marcin Choromanski}, \bibinfo{person}{Adrian Wong}, \bibinfo{person}{Stefan Welker}, \bibinfo{person}{Federico Tombari}, \bibinfo{person}{Aveek Purohit}, \bibinfo{person}{Michael~S Ryoo}, \bibinfo{person}{Vikas Sindhwani}, \bibinfo{person}{Johnny Lee}, {et~al\mbox{.}}} \bibinfo{year}{2022}\natexlab{}.
\newblock \showarticletitle{Socratic Models: Composing Zero-Shot Multimodal Reasoning with Language}. In \bibinfo{booktitle}{\emph{The Eleventh International Conference on Learning Representations}}.
\newblock


\bibitem[Zhai et~al\mbox{.}(2021)]%
        {zhai2021lit}
\bibfield{author}{\bibinfo{person}{Xiaohua Zhai}, \bibinfo{person}{Xiao Wang}, \bibinfo{person}{Basil Mustafa}, \bibinfo{person}{Andreas Steiner}, \bibinfo{person}{Daniel Keysers}, \bibinfo{person}{Alexander Kolesnikov}, {and} \bibinfo{person}{Lucas Beyer}.} \bibinfo{year}{2021}\natexlab{}.
\newblock \showarticletitle{LiT: Zero-Shot Transfer with Locked-image Text Tuning}.
\newblock \bibinfo{journal}{\emph{arXiv preprint arXiv:2111.07991}} (\bibinfo{year}{2021}).
\newblock


\bibitem[Zhang et~al\mbox{.}(2022)]%
        {zhang2022a}
\bibfield{author}{\bibinfo{person}{Ruohan Zhang}, \bibinfo{person}{Dhruva Bansal}, \bibinfo{person}{Yilun Hao}, \bibinfo{person}{Ayano Hiranaka}, \bibinfo{person}{Jialu Gao}, \bibinfo{person}{Chen Wang}, \bibinfo{person}{Roberto Mart{\'\i}n-Mart{\'\i}n}, \bibinfo{person}{Li Fei-Fei}, {and} \bibinfo{person}{Jiajun Wu}.} \bibinfo{year}{2022}\natexlab{}.
\newblock \showarticletitle{A Dual Representation Framework for Robot Learning with Human Guidance}. In \bibinfo{booktitle}{\emph{6th Annual Conference on Robot Learning}}.
\newblock
\urldef\tempurl%
\url{https://openreview.net/forum?id=H6rr_CGzV9y}
\showURL{%
\tempurl}


\end{thebibliography}
